\documentclass{article}
\usepackage[preprint]{colm2026_conference}
\usepackage{fvextra}

\usepackage{amsmath,amsfonts,bm}

\def\eqref#1{equation~\ref{#1}}

\def\1{\bm{1}}

\DeclareMathAlphabet{\mathsfit}{\encodingdefault}{\sfdefault}{m}{sl}
\SetMathAlphabet{\mathsfit}{bold}{\encodingdefault}{\sfdefault}{bx}{n}

\usepackage[T1]{fontenc}
\usepackage{multirow}
\usepackage{algpseudocode}
\usepackage{algorithm}
\usepackage{algpseudocode}
\usepackage{nicefrac}
\usepackage{booktabs}
\usepackage{makecell}
\usepackage{placeins}
\usepackage{hyperref}
\usepackage{xcolor}
\usepackage{url}
\usepackage{enumitem}
\usepackage{bbding}
\usepackage{pifont}
\usepackage{varwidth}
\usepackage[most]{tcolorbox}
\usepackage{booktabs}
\usepackage{siunitx}
\usepackage[table]{xcolor}
\usepackage{colortbl}
\usepackage{float}
\usepackage{tabularx}
\usepackage{threeparttable}
\usepackage{url}
\usepackage{lineno}
\usepackage{wrapfig}
\usepackage{needspace}
\usepackage{microtype}
\usepackage{caption}

\definecolor{darkblue}{rgb}{0, 0, 0.5}
\hypersetup{colorlinks=true, citecolor=darkblue, linkcolor=darkblue, urlcolor=darkblue}
\usepackage{graphicx}

\title{BioAlchemy: Distilling Biological Literature into Reasoning-Ready Reinforcement Learning Training Data}

\author{
  \parbox{0.85\textwidth}{
    Brian Hsu$^{1,2}$,
    Ozan Gökdemir$^{1,2}$,
    Carlo Siebenschuh$^{1,2}$,
    Bruce Parrello$^{1,2}$,
    Neil Getty$^{2}$,
    Thomas S. Brettin$^{2}$,
    Rick L. Stevens$^{1,2}$,
    Ian T. Foster$^{1,2}$,
    Nicholas Chia$^{2,*}$,
    Arvind Ramanathan$^{2,*}$
  }
  \\[1.5em]
  \normalfont
  $^1$Department of Computer Science, University of Chicago, Chicago, IL 60637 \\
  $^2$Data Science and Learning Division, Argonne National Laboratory, Lemont, IL 60439 \\[0.3em]
  {\small \texttt{bah4228@uchicago.edu}, \texttt{\{chia, ramanathana\}@anl.gov}} \\[0.3em]
  {\small $^*$Equal contribution}
}
\usepackage{xcolor}

\begin{document}

\ifcolmsubmission
\linenumbers
\fi

\begingroup
\microtypesetup{protrusion=false, expansion=false}
\maketitle
\endgroup

\begin{abstract}
Despite the large corpus of biology training text, the impact of reasoning models on biological research generally lags behind math and coding. In this work, we show that biology questions from current large-scale reasoning datasets do not align well with modern research topic distributions in biology, and that this topic imbalance may negatively affect performance. In addition, we find that methods for extracting challenging and verifiable research problems from biology research text are a critical yet underdeveloped ingredient in applying reinforcement learning for better performance on biology research tasks. We introduce BioAlchemy, a pipeline for sourcing a diverse set of verifiable question-and-answer pairs from a scientific corpus of biology research text. We curate BioAlchemy-345K, a training dataset containing over 345K scientific reasoning problems in biology. Then, we demonstrate how aligning our dataset to the topic distribution of modern scientific biology can be used with reinforcement learning to improve reasoning performance. Finally, we present BioAlchemist-8B, which improves over its base reasoning model by 9.12\% on biology benchmarks. These results demonstrate the efficacy of our approach for developing stronger scientific reasoning capabilities in biology. The BioAlchemist-8B model is available at: \url{https://huggingface.co/BioAlchemy}.

\end{abstract}

\section{Introduction}

Reasoning models have demonstrated the ability to mimic human chain-of-thought processes as a means to improve on tasks requiring complex reasoning. This has been dramatically demonstrated with gold medal performances from multiple reasoning models at the International Math Olympiad in 2025 \citep{huang2025gemini}. Similarly difficult tests in other areas of science have shown a remarkable gap between math or code and the other sciences \citep{zhu2025probing}. This has led to the development of datasets such as NaturalReasoning \citep{naturalreasoning}, TextbookReasoning \citep{megascience}, and Nemotron-Science-v1 \citep{nemotron_science_v1}
for improving performance across the sciences more broadly.

\begin{wrapfigure}{tr}{0.55\textwidth}
  \centering
  \vspace{-0.8\baselineskip}
  \includegraphics[width=0.94\linewidth,trim=0 0 0 20,clip]{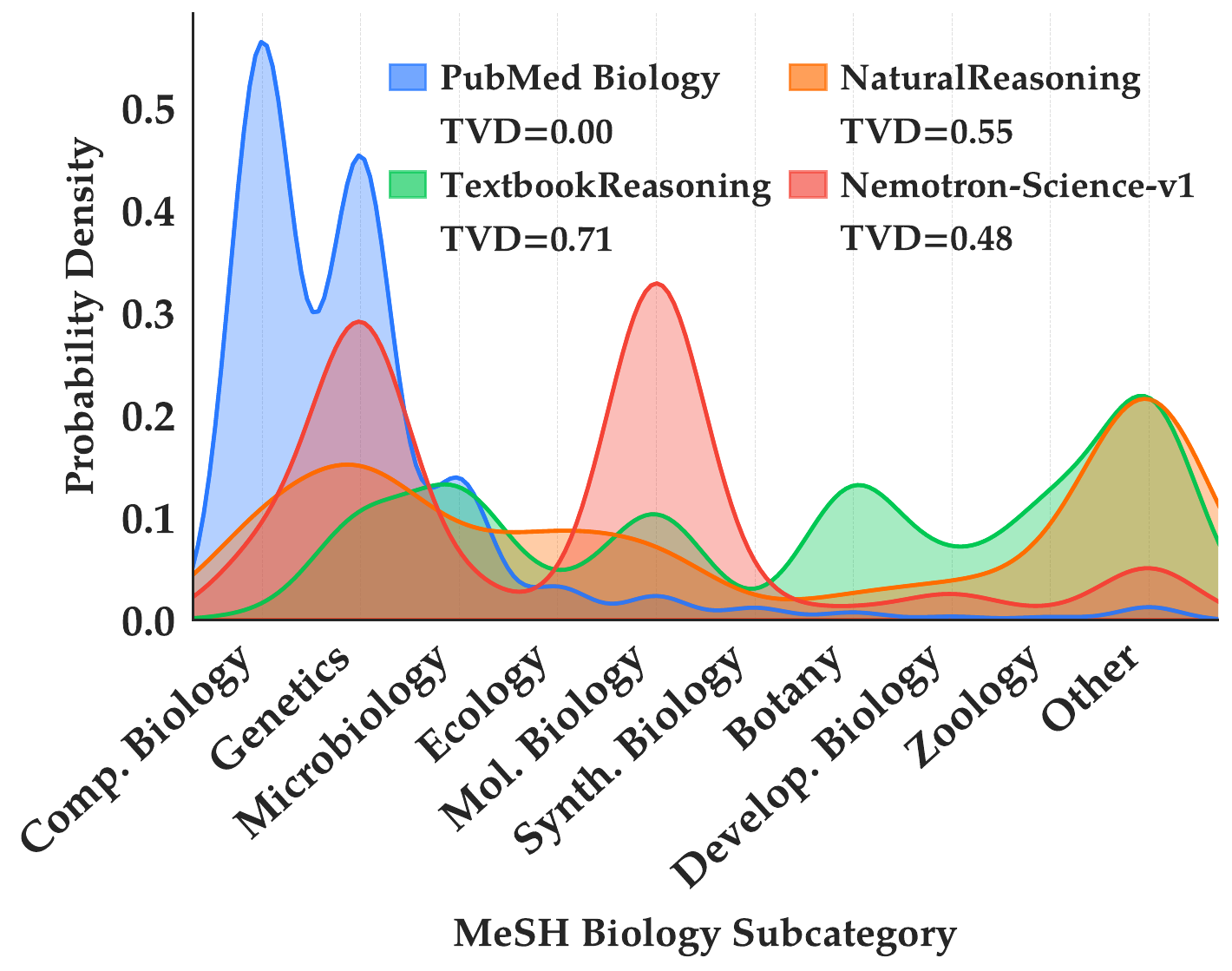}
  \caption{Content distribution differences between reasoning problems in biology. Comparison is based on MeSH topics representing PubMed research articles from 2020--2024.}
  \label{fig:kde-comparison-ex-bioalchemy}
  \vspace{-0.8\baselineskip}
\end{wrapfigure}

Biology is a particularly challenging area for reasoning models due to the number of disparate and difficult subtasks within the biological thinking process -- a process that couples databases worth of factual recall with contextually-dependent mechanistic insight. Benchmarks such as LAB-Bench \citep{labbench} and GPQA-Diamond \citep{rein2024gpqa} demonstrate this breadth, spanning database search, sequence manipulation, and molecular biology. Worse, explicit reasoning about biological concepts differ semantically and significantly from heterogeneous research texts that detail experimental setup, environmental context, and qualitative observations. In other words, reasoning for biology is not directly represented in current training corpora to the same extent that it is for math or code. Yet, training data across reasoning problems in biology remains heavily skewed. Figure~\ref{fig:kde-comparison-ex-bioalchemy} presents Medical Subject Headings (MeSH) topic distributions computed over biology questions from some of the largest scientific reasoning datasets derived from pretraining and textbook corpora, where MeSH serves as a standardized vocabulary for organizing topics in biomedical literature \citep{mesh_biology}. The biology reasoning problems in these datasets are skewed away from the distribution of biology literature. This means that any potential set of knowledge or skills being learned cannot be expected to reasonably reflect modern biology research needs. This can be troublesome as the topic distributions neither reflect the relative importance of modern biology topic areas nor help smooth the distribution noticeably (Table~\ref{tab:topic-distributions-full}).

In this work, we investigate how reasoning problems aligned at the subtopic level for biology can be used to drive gains in biological reasoning. We use an LLM-driven pipeline to source over 345K reasoning problems for reinforcement learning (RL) that capture subtopics and relationships representative of modern scientific biology. Through topic sampling techniques, we derive a reasoning dataset that reflects modern scientific biology at the subtopic level. Through various experiments, we demonstrate the utility of our reasoning dataset in eliciting performance across various biology benchmarks. We also provide an analysis on which composition of reasoning topics can lead to the best gains in performance. Our work provides a dataset and strategy for how to use topic-level alignment to train reasoning models that can better solve challenges in scientific biology.

\section{Background and related work}

Reasoning models have recently achieved state-of-the-art performance on various math and coding benchmarks \citep{shao2024deepseekmath}. A driving factor in this rise in performance is an abundance of math and coding reasoning datasets such as OpenMathReasoning \citep{openmathreasoning} and NuminaMath \citep{numina_math_datasets}. However, translating the same results to scientific domains has remained an open challenge due to the relative sparsity in such datasets compared to math and coding. 

Recent works such as MegaScience \citep{megascience}, NaturalReasoning \citep{naturalreasoning}, and Nemotron-Science-v1 \citep{nemotron_science_v1} have attempted to bridge this gap through large reasoning datasets covering broad topics such as biology, chemistry, math, etc. However, these works do not use scientific research papers as the core data source for deriving reasoning problems, instead using pretraining corpora, textbooks, or synthetically generated questions. Furthermore, these datasets attempt to improve biological reasoning through generalized gains in scientific reasoning, without sufficient attention on the topic complexities inherent to biology. Other efforts have focused on synthesizing data that aligns well for biological reasoning. BioReason \citep{fallahpour2025bioreason} introduces multi-modal data that aligns genomic sequences with textual reasoning. PaperSearchQA \citep{papersearchqa} curates reasoning problems and a retrieval corpus from PubMed abstracts for reinforcement learning. However, these works do not consider the distribution of biological reasoning problems at the subtopic level. \citet{blakeney2024does} and \citet{mizrahi2025language} have argued for proper upsampling and calibration of domain data during training for downstream performance in relevant benchmarks. Therefore, we propose that in order to achieve better performance in biological research, one should curate reasoning data that reflects modern biological research topics on a more granular level, motivating our work.

\section{Dataset: BioAlchemy-345K}\label{gen_inst}

The BioAlchemy dataset is built from a scientific corpus sourced from the American Society for Microbiology (ASM) \citep{ASM}, PLoS Genetics \citep{plos_genetics_journal}, PLoS Computational Biology \citep{plos_compbio_journal}, and Semantic Scholar \citep{semanticscholar}. We use an automated question-answer (QA) generation pipeline similar to \citet{AutomatedMCQA} that combines a distributed parser \citep{adaparse} with high-throughput chunking strategies \citep{HiPerRAG}. Our QA generation process is applied over scientific text to generate a provenance-driven, scientifically-grounded, and predominantly MCQA dataset that is appropriate for RL training. We use multiple prompts for deriving our QA pairs, which are detailed in Section~\ref{sec:qa-generation-prompts}. In total, our aggregate dataset is sourced from 77.7K scientific research articles and contains 345K reasoning problems appropriate for RL. 

\begin{wraptable}{r}{0.55\textwidth}
\vspace{-0.6\baselineskip}
\centering
\small
\setlength{\tabcolsep}{3pt}
\renewcommand{\arraystretch}{0.95}

\begin{tabularx}{\linewidth}{@{}Xcc@{}}
\toprule
\textbf{Data Source} & 
\textbf{\# Research Articles} & 
\textbf{\# QA Pairs} 
\\
\midrule
ASM & 35.8K & 173.6K \\
PLoS CompBio & 10.5K & 76.1K \\
PLoS Genetics & 8.9K & 56.9K \\
Semantic Scholar & 22.5K & 38.4K \\
\midrule
\textbf{Total} & 
\textbf{77.7K} & 
\textbf{345K}
\\
\bottomrule
\end{tabularx}

\caption{Composition of dataset for BioAlchemy.}
\label{fig:data-source-counts}
\vspace{-0.8\baselineskip}
\end{wraptable}

Table~\ref{tab:dataset-qualitative-comparison} shows that BioAlchemy compares favorably to the biology subsets of other large-scale reasoning datasets along the dimensions most relevant for reinforcement learning. To make this comparison, we extract the biology problems from the relevant datasets through a multi-label topic classifier, described in Section~\ref{sec:sampling-topic-distributions}. Among the datasets considered, BioAlchemy contains the largest number of biology reasoning questions, with 345K examples, making it 6.5$\times$ larger than the biology subset for the next largest dataset, NaturalReasoning. In addition to scale, BioAlchemy is almost entirely composed of reasoning problems with ground truth answers that are easily verifiable via exact-match, making it especially well-suited for reinforcement learning with verifiable rewards \citep{wen2025reinforcementlearningverifiablerewards}. The source of reasoning problems also varies considerably across datasets. NaturalReasoning extracts questions from pretraining web corpora such as DCLM-Baseline, which are broad in coverage but not curated for scientific depth. TextbookReasoning from MegaScience derives its questions from university textbooks, which are more structured but reflect pedagogical rather than research-level content. Nemotron-Science-v1 generates its questions synthetically via LLMs, without grounding in primary scientific literature \citep{nemotron_science_v1}. In contrast, BioAlchemy is sourced directly from scientific research publications, grounding its reasoning problems in the language, concepts, and evidence structures of active biological research. It is worth noting that our design intention behind grounding our verifiable reasoning problems in scientific literature is to better preserve the ability to expand model knowledge without solely relying on distillation from state-of-the-art LLMs. Smaller models typically exhibit strong gains from distillation \citep{deepseek}, and the quality of training data may matter more at larger models where capacity constraints are reduced \citep{zhang-etal-2025-capacitygap}. Since we are limited to an 8B model, we leave this as an open research question for future work.

\begin{table}[!htp]
\centering
\small
\setlength{\tabcolsep}{6pt}
\renewcommand{\arraystretch}{1.20}

\resizebox{\linewidth}{!}{%
\begin{tabular}{lccccc}
\toprule
\textbf{Dataset} &
\textbf{Source} &
\textbf{Size (Biology)} & 
\textbf{Training} & 
\textbf{MeSH Coverage} \\
\midrule
BioAlchemy
& Scientific Research Text 
& \textbf{345K} 
& RL (GRPO) 
& \textbf{95.6\%} \\
NaturalReasoning
& Pretraining Corpora 
& 52.9K
& SFT 
& 34.7\% \\
TextbookReasoning
& University Textbooks 
& 51.2K
& SFT 
& 66.6\% \\
Nemotron-Science-v1
& N/A 
& 29.1K
& RL (GRPO) 
& 91.4\% \\
\bottomrule
\end{tabular}%
}
\caption{Comparison of reasoning datasets, where dataset sizes are based on questions classified as biology questions. BioAlchemy is the largest and has the strongest coverage.}
\label{tab:dataset-qualitative-comparison}
\end{table}

\subsection{Sampling topic distributions}\label{sec:sampling-topic-distributions}

Previous work has emphasized the importance of aligning training distributions for optimal domain performance \citep{gururangan-etal-2020-dont,chawla2025quantifying}. Therefore, we choose to align the distribution of our reasoning dataset with PubMed articles published from 2020 to 2024 at the MeSH Biology subtopic level \citep{pubmed}. Specifically, given the 345K total reasoning problems sourced via BioAlchemy, which we denote as $\mathcal{D}$, we sample a reasoning dataset $\mathcal{S} \subseteq \mathcal{D}$ such that its categorical distribution $P_{\mathcal{S}}$ approximates $P_\text{Bio}$, which represents the topic distribution capturing modern biology research text. The alignment process for creating a calibrated dataset $\mathcal{S}$ provides a greater level of granular control over the reasoning datasets we compare to in this work on a subtopic level.

\begin{algorithm}[!tp]
\caption{Greedy Removal for Distribution Matching}
\label{alg:greedy-removal}
\begin{algorithmic}[1]
\Require Dataset $\mathcal{D}$, target distribution $\hat{P}_\text{Bio}$, threshold $\tau$, penalty $\lambda$
\State $\mathcal{S} \leftarrow \{i \in \mathcal{D} : |C_i| > 0\}$ \Comment{Initialize with all valid samples}
\State Compute category counts $n_j = \sum_{i \in \mathcal{S}} \mathbf{1}[c_j \in C_i]$
\While{$\mathrm{TVD}(\hat{P}_\mathcal{S}, \hat{P}_\text{Bio}) > \tau$ \textbf{and} $|\mathcal{S}| > n_{\min}$}
    \State Identify overrepresented categories: $\mathcal{O} = \{c_j : \hat{p}_{c_j} > \hat{p}_{c_j}^{\text{Bio}}\}$
    \For{each sample $i \in \mathcal{S}$}
        \State $\mathrm{score}(i) \leftarrow
        \sum_{c_j \in C_i \cap \mathcal{O}} (\hat{p}_{c_j} - \hat{p}_{c_j}^\text{Bio})
        - \lambda \sum_{c_j \in C_i \setminus \mathcal{O}} (\hat{p}_{c_j}^{\text{Bio}} - \hat{p}_{c_j})$
    \EndFor
    \State Remove the top-k highest-scoring samples from $\mathcal{S}$
    \State Update counts $n_j$ and recompute $\hat{P}_\mathcal{S}$
\EndWhile
\Statex
\Return $\mathcal{S}$
\end{algorithmic}
\end{algorithm}

To approximate $P_\text{Bio}$, we use the NCBI Entrez API \citep{sayers2009eutilities} to collect a set of scientific publications $\mathcal{Q}$. For each document $d \in \mathcal{Q}$, we download its associated set of MeSH topics $M_{d} \subseteq \mathcal{C}$, where $\mathcal{C}=\{c_{1}, \ldots, c_{21}\}$ is the canonical support set of 21 MeSH Biology subcategories under the Biology hierarchy H01.158.273.* \citep{mesh_biology}. Using our sampled documents and their associated MeSH category sets, we compute the empirical distribution of MeSH topic categories $\hat{P}_{\text{Bio}}$ over sampled documents:
\begin{equation}
\hat{P}_{\text{Bio}} =
(\hat{p}^{\text{Bio}}_{c_{1}},\, \hat{p}^{\text{Bio}}_{c_{2}},\, \ldots,\, \hat{p}^{\text{Bio}}_{c_{21}}),
\;\text{where}\;
\hat{p}^{\text{Bio}}_{c_{j}} = \frac{\sum_{d \in \mathcal{Q}} \mathbf{1}[c_{j} \in M_{d}]}{|Q|},
\; j \in [21]
\end{equation}

\begin{wrapfigure}{r}{0.43\textwidth}
  \centering
  \includegraphics[width=\linewidth,
  trim=2 5 0 5,clip]{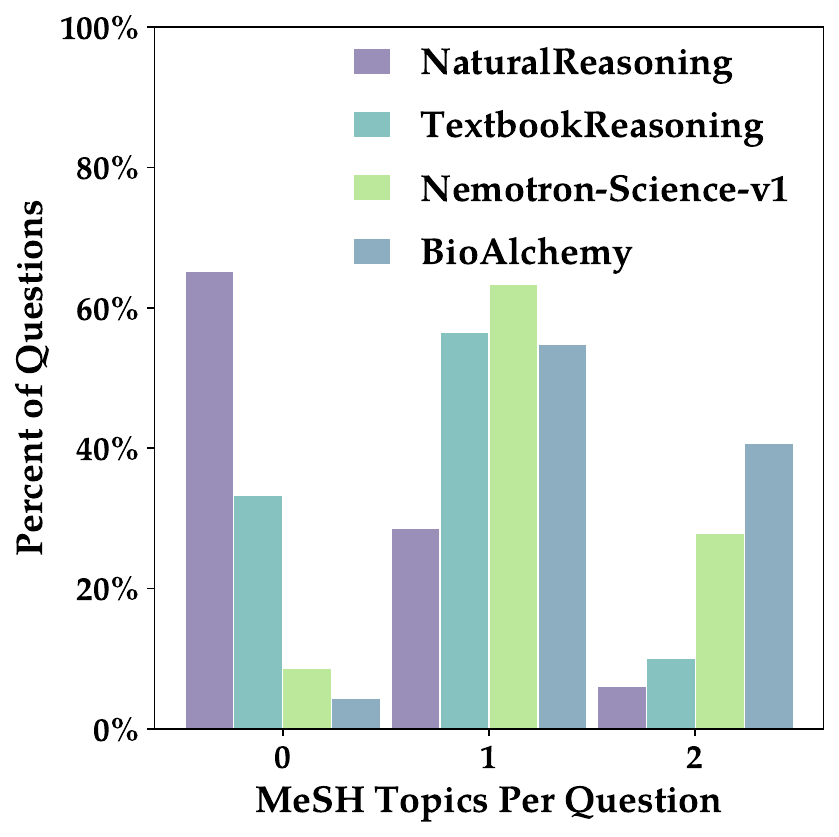}
  \caption{BioAlchemy has more MeSH topics per biology question. Comparison is based on 50K subsamples.}
  \label{fig:topic-count-histogram}
\end{wrapfigure}

To compute $\hat{P}_{\mathcal{S}}$, we first have to train a multi-label classifier to label each reasoning problem from our dataset $\mathcal{D}$ with its relevant set of MeSH Biology subcategory labels. To do this, we first collect an additional set of papers and MeSH topic array labels via NCBI API for our training and validation sets. We then prompt-tuned a topic classifier on these sets using DSPy \citep{khattab2023dspy}, with GPT-4.1 as the underlying model. We select few-shot examples via a bootstrap with random search over random prompt configurations, and evaluate 11 candidate prompt configurations against our stratified validation set using micro-averaged F1 (Table~\ref{tab:mesh_training_data}). Our optimized classifier achieves 72.7\% F1, a +96\% relative improvement over the zero-shot baseline (37.0\% F1). Compared to single-label supervision, using multi-label ground truth also yields an additional +23.4\% gain. We now label each reasoning problem $i\in\mathcal{D}$ in our dataset by classifying it with a set of subcategory labels $C_i \subseteq \mathcal{C}$ that describes the content of the reasoning problem. Note that we allow for $C_i=\emptyset$, as certain reasoning problems may not contain content relevant to MeSH Biology subtopics despite being labeled as biology. This may happen due to the natural adjacency between medical and biology topics, as well as erroneous safety filtering from our LLM API, based on manual inspection of a sample of filtered questions. Using our classification method, we find that our reasoning problems are more diverse than other datasets in MeSH content, as demonstrated in Figure~\ref{fig:topic-count-histogram}.

We now construct our reasoning dataset $\mathcal{S} = \{i \in \mathcal{D} : \mathrm{TVD}(\hat{P}_{\mathcal{S}}, \hat{P}_{\text{Bio}}) \leq \tau \}$. We set $\tau = 0.21$ based on our classifier's prediction error; since our classifier has an error rate of $\epsilon = 0.273$, the maximum TVD perturbation attributable to misclassification is bounded by $\epsilon \cdot \mathrm{TVD}(P_{\text{Bio}}, U) = 0.213$, where $U$ is the uniform distribution. Distributional differences below this threshold are therefore indistinguishable from classification noise, making $\tau=0.21$ the tightest meaningful target. To construct $\mathcal{S}$, we apply a greedy sampling algorithm on $\mathcal{D}$ that iteratively removes reasoning problems that contribute the most to $\mathrm{TVD}(\hat{P}_{\mathcal{S}}, \hat{P}_{\text{Bio}})$, as shown in Algorithm~\ref{alg:greedy-removal}. We score each sample as a weighted estimate of its deviation from the target distribution, prioritizing keeping samples that contain under-represented MeSH categories, since certain categories like Synthetic Biology are extremely rare. We then remove samples with the highest score, reducing the over-representation of categories that dominate the source distribution.

\begin{table}[!htbp]
\centering
\resizebox{\linewidth}{!}{%
\begin{tabular}{clrrrrr}
\toprule
\multicolumn{1}{c}{\textbf{Rank}} &
\multicolumn{1}{c}{\textbf{MeSH Biology Subcategory}} &
\multicolumn{1}{c}{\makecell{\textbf{PubMed}\\\textbf{2020--2024}}} &
\multicolumn{1}{c}{\makecell{\textbf{Textbook}\\\textbf{Reasoning}}} &
\multicolumn{1}{c}{\makecell{\textbf{Natural}\\\textbf{Reasoning}}} &
\multicolumn{1}{c}{\makecell{\textbf{Nemotron}\\\textbf{Science-v1}}} &
\multicolumn{1}{c}{\makecell{\textbf{BioAlchemy}}} \\
\midrule
1  & Computational Biology     & 45.32\% & 0.57\%  & 12.09\% & 7.44\%  & 24.95\% \\
2  & Genetics                  & 36.18\% & 11.37\% & 18.30\% & 34.67\% & 36.89\% \\
3  & Microbiology              & 10.90\% & 14.39\% & 7.19\%  & 3.78\%  & 16.75\% \\
4  & Ecology                   & 2.56\%  & 2.51\%  & 9.24\%  & 0.99\%  & 2.80\% \\
5  & Molecular Biology         & 1.88\%  & 12.59\% & 8.45\%  & 40.41\% & 6.63\% \\
6  & Synthetic Biology         & 0.97\%  & 0.04\%  & 0.63\%  & 1.76\%  & 0.50\% \\
7  & Botany                    & 0.61\%  & 15.72\% & 2.86\%  & 1.13\%  & 0.79\% \\
8  & Developmental Biology     & 0.29\%  & 5.36\%  & 3.84\%  & 2.94\%  & 2.71\% \\
9  & Zoology                   & 0.26\%  & 11.86\% & 4.43\%  & 0.67\%  & 0.31\% \\
10 & Other                     & 1.03\%  & 25.59\% & 32.98\% & 6.21\%  & 7.67\% \\
\midrule
   & \textbf{TVD}              & \textbf{N/A} & \textbf{0.71} & \textbf{0.55} & \textbf{0.48} & \textbf{0.21} \\
\bottomrule
\end{tabular}%
}
\caption{BioAlchemy has better distributional alignment at the subtopic level compared to the biology subsets of TextbookReasoning, NaturalReasoning, and Nemotron-Science-v1. Percentages are not mutually exclusive as publications/QA pairs can be indexed under multiple MeSH terms.}
\label{tab:mesh_comparison_small}
\end{table}

As seen in Table~\ref{tab:mesh_comparison_small}, the best-calibrated reasoning dataset we sample is more closely aligned, at the topic level, with the distribution of scientific publications drawn from PubMed compared to the biology questions presented in NaturalReasoning \citep{naturalreasoning}, TextbookReasoning \citep{megascience}, and Nemotron-Science-v1 \citep{nemotron_science_v1}. We find that TextbookReasoning and NaturalReasoning have reasoning problems with MeSH topics belonging to less represented categories, such as Zoology. This might reflect inherent distributional differences in source material used by MegaScience and NaturalReasoning. 

\begin{table}[!htbp]
\centering
\small
\setlength{\tabcolsep}{6pt}
\renewcommand{\arraystretch}{1.15}
\begin{tabular}{lccc}
\toprule
\textbf{Data Source} 
& \makecell{\textbf{Mean }\\\textbf{MeSH Topics}} 
& \makecell{\textbf{\% Non-Zero}\\\textbf{MeSH Topics}}
& \makecell{\textbf{\% With $\geq 2$}\\\textbf{MeSH Topics}} \\
\midrule
NaturalReasoning
& 0.41  
& 34.7\%  
& 6.2\%  \\
TextbookReasoning  
& 0.77  
& 66.6\%  
& 10.1\%  \\
Nemotron-Science-v1       
& 1.19 
& 91.4\% 
& 28.0\% \\
BioAlchemy        
& \textbf{1.37} 
& \textbf{95.6\%} 
& \textbf{40.8\%} \\
\bottomrule
\end{tabular}
\caption{BioAlchemy has higher MeSH topic coverage and higher average number of topics. The average number of MeSH subcategories for PubMed publications is 1.07.}
\label{tab:mesh-topic-coverage}
\end{table}

In addition to having better distributional alignment, BioAlchemy also has broader coverage and density of MeSH content per reasoning question. Seen in Figure~\ref{fig:topic-count-histogram}, BioAlchemy has a much higher percentage of reasoning questions that contain at least one MeSH topic, suggesting that it has better coverage in scientific biology concepts. Furthermore, as seen in Table~\ref{tab:mesh-topic-coverage}, BioAlchemy also contains more MeSH topics on average than both TextbookReasoning, NaturalReasoning, and Nemotron-Science-v1, reflecting the interdisciplinary aspects of current biological research.

\subsection{Exponential smoothing}\label{sec:exponential-smoothing}

While our sampling algorithm in Section~\ref{sec:sampling-topic-distributions} successfully aligns the reasoning dataset with the PubMed topic distribution, PubMed itself exhibits substantial class imbalance across MeSH Biology subcategories, as seen in Table~\ref{tab:mesh_comparison_small}. To mitigate this inherited imbalance during training, we apply exponential smoothing to our sampled distribution $P_{\mathcal{S}}$ following language upsampling strategies widely used in multilingual language model pretraining. 

To do this, we first compute smoothed versions of the PubMed topic distribution  $P^{\alpha}_{\text{Bio}}$, where decreasing $\alpha$ applies stronger upsampling of the tail categories to make the distribution more uniform. Then, we apply Algorithm~\ref{alg:greedy-removal} to $\mathcal{D}$ to generate exponentially smoothed datasets $\mathcal{S}_{\alpha}$ that each align with a variation of $P^{\alpha}_{\text{Bio}}$. To compute each $P^{\alpha}_{\text{Bio}}$, the sampling probability for each subcategory $c_{j}$ is computed as ${p}^{\alpha}_{c_j} \propto n_j^{\alpha}$, where $n_j$ is the publication count and
$\alpha \in (0,1]$ controls the degree of smoothing -- a formulation
originally developed to prevent high-resource languages from drowning out
low-resource ones in multilingual pretraining. Multilingual BERT
\citep{devlin2019bert} used $\alpha = 0.7$ to balance high- and low-resource
languages during pretraining on 104 Wikipedia corpora; the same value was
subsequently adopted by mBART for 25-language denoising pretraining
\citep{liu2020multilingual}. Later work at larger scale found more aggressive
smoothing to be beneficial: both XLM-R \citep{conneau2020unsupervised} and mT5
\citep{xue2021mt5} independently
converged on $\alpha = 0.3$ as optimal for overall cross-lingual transfer
performance. For most of the experiments in this paper, we chose the midpoint of these prior studies, $\alpha = 0.5$.

\begin{table}[!htp]
\centering
\small
\setlength{\tabcolsep}{6pt}
\renewcommand{\arraystretch}{1.15}

\resizebox{\textwidth}{!}{%
\begin{threeparttable}
\begin{tabular}{lccccc}
\toprule
\textbf{Dataset}
& \textbf{Size}
& \textbf{Topic Max/Min Ratio}
& \textbf{Avg. Upsample$^{\dagger}$}
& \textbf{Sample Yield}
& \textbf{TVD}
\\
\midrule
Full Dataset $\mathcal{D}$
& 345K
& 9,301$\times$
& 11.8$\times$
& 100\%
& 0.52
\\
\midrule
Dataset $\mathcal{S}_{\alpha=1.0}$
& 126.7K
& 7,932$\times$
& 1.0$\times$
& 36.73\%
& 0.21
\\
Dataset $\mathcal{S}_{\alpha=0.7}$
& 146.2K
& 536$\times$
& 9.7$\times$
& 42.38\%
& 0.25
\\
Dataset $\mathcal{S}_{\alpha=0.5}$
& 152.0K
& 89$\times$
& 39.6$\times$
& 44.06\%
& 0.27
\\
Dataset $\mathcal{S}_{\alpha=0.3}$
& 83.2K
& 15$\times$
& 134.3$\times$
& 24.10\%
& 0.39
\\
\bottomrule
\end{tabular}

\begin{tablenotes}[flushleft]
  \small
  \item[$\dagger$] Avg. upsample is mean ratio of sampling probabilities ($\nicefrac{p_{c_{j}}^{\alpha}}{p_{c_{j}}^{\mathrm{orig}}}$) from the 5 rarest categories in \hspace*{0.6em}PubMed~2020--2024. See Table~\ref{tab:smoothed_sampling_full} for full counts.
\end{tablenotes}

\caption{Sub-sampling BioAlchemy-345K with increasing representation of low-frequency topics from exponential smoothing. $\mathcal{S}_{\alpha=0.5}$ provides the best tradeoff in terms of dataset
size, distributional alignment, and upsampling of rare topics.}

\label{tab:alpha-sample-summary}
\end{threeparttable}
}%
\end{table}

\section{Experimental setup}

\subsection{Training}

We use two training regimes for our experiments: reinforcement learning with verifiable rewards \citep{wen2025reinforcementlearningverifiablerewards}, and supervised fine-tuning \citep{ouyang2022training}. We use outcome-based Group Relative Policy Optimization (GRPO) with verifiable rewards, which allows us to estimate advantages without a parameterized critic model \citep{deepseek}. Our reward is defined on a scale of $r \in [0, 2]$, which includes small rewards for formatting and small penalties for repeated answers. To stabilize response lengths and promote exploration, we use a token-level loss and higher clipping of the policy gradient \citep{dapo}, while removing normalization to avoid question-level difficulty bias \citep{liu2025understanding}. Each experiment uses its own data and training configuration, allowing us to test robustness across different settings (Sections~\ref{sec:appendix-dataset-comparison},~\ref{sec:appendix-distribution-comparison}). We use the VeRL framework for training GRPO and SFT \citep{verl_github}, and train on Qwen3-8B \citep{yang2025qwen3}. 

\subsection{Evaluations}

To test the utility of our dataset in eliciting scientific reasoning in biology, we evaluate our models on the SeqQA, ProtocolQA, and Cloning Scenarios subtasks from LAB-Bench \citep{labbench}, which test model capabilities in understanding molecular biology workflows, protocol reasoning, and real-world cloning scenarios. Performance along these axes can help gauge the potential utility of our model in helping design more intelligent bio-protocols and workflows, which has had rising interest in recent years \citep{liu2025bioprobench, odonoghue2023bioplanner}. We also test our model on the full 1K PQA-L task in PubMedQA \citep{jin2019pubmedqa} to evaluate our model on biomedical understanding. Finally, we evaluate our models on the biology subtasks in GPQA-Diamond, Genetics and Molecular Biology \citep{rein2024gpqa}, and compute a general average. On LAB-Bench tasks, we base our evaluation harness on the repository given by FutureHouse \citep{futurehouse_labbench_repo_2025}. For PubMedQA and GPQA-Diamond, we use our own evaluation harness, using exact-match for computing accuracy. We report the average accuracies and confidence intervals averaged over 10 independent runs to quantify run-to-run variability (Table~\ref{tab:inference-settings}).

\section{Experiments}\label{sec:experiments}
\label{headings}

\subsection{Dataset performance}\label{sec:sftvsrl}

We evaluate the utility of our dataset for both SFT and RL. For SFT, we subsample reasoning problems from BioAlchemy and collect one-shot responses from Llama-3.3 70B Instruct as reasoning traces. NaturalReasoning and TextbookReasoning each contain around 50K biology reasoning problems, whereas Nemotron-Science-v1 contains around 29K. To keep training scale comparable, we train on 50K SFT examples drawn from the BioAlchemy split $\mathcal{S}_{\alpha=0.5}$. We do not run RL with verifiable rewards on NaturalReasoning and TextbookReasoning, as we find that their biology problems are not easily converted into a format that is appropriate for RL with verifiable rewards. 

\FloatBarrier
\begin{table*}[!htbp]
\centering
\small
\setlength{\tabcolsep}{5pt}
\renewcommand{\arraystretch}{1.12}
\resizebox{\textwidth}{!}{%
\begin{tabular}{lccccc}
\toprule
\textbf{Benchmark}
& \multicolumn{3}{c}{\textbf{SFT}}
& \multicolumn{2}{c}{\textbf{RL}} \\
\cmidrule(lr){2-4}\cmidrule(lr){5-6}
& \makecell{\textbf{Natural}\\\textbf{Reasoning}}& \makecell{\textbf{Textbook}\\\textbf{Reasoning}}& \makecell{\textbf{BioAlchemy}}& \makecell{\textbf{Nemotron-}\\\textbf{Science-v1}}& \makecell{\textbf{BioAlchemy}} \\
\midrule
\textbf{ProtocolQA} 
& 43.61$\pm$2.55\% 
& 43.33$\pm$2.89\% 
& 43.80$\pm$2.10\% 
& 44.44$\pm$2.09\% 
& \textbf{45.83$\pm$2.32\%} \\
\textbf{SeqQA} 
& 10.43$\pm$0.75\% 
& 15.20$\pm$0.73\% 
& 14.03$\pm$0.69\% 
& \textbf{20.42$\pm$0.66\%} 
& 18.73$\pm$0.85\% \\
\textbf{Cloning Scenarios} 
& 8.48$\pm$1.99\% 
& 8.18$\pm$2.30\% 
& 9.70$\pm$2.85\% 
& \textbf{13.94$\pm$3.85\%} 
& 11.52$\pm$3.03\% \\
\midrule
\textbf{PubMedQA} 
& 67.89$\pm$0.60\% 
& \textbf{72.47$\pm$0.15\%} 
& 69.44$\pm$0.24\% 
& 69.21$\pm$0.38\% 
& 68.09$\pm$0.28\% \\
\textbf{GPQA-Bio} 
& 36.84$\pm$0.00\% 
& 42.11$\pm$0.00\% 
& 44.21$\pm$3.64\% 
& 47.89$\pm$1.19\% 
& \textbf{57.89$\pm$0.00\%} \\
\midrule
\textbf{Overall Avg.} 
& 33.45$\pm$0.67\% 
& 36.26$\pm$0.75\% 
& 36.24$\pm$1.03\% 
& 39.18$\pm$0.92\% 
& \textbf{40.41$\pm$0.78\%} \\
\bottomrule
\end{tabular}%
}
\caption{Comparison of reasoning datasets across relevant training strategies. BioAlchemy with GRPO performs best overall.}
\label{tab:dataset-benchmark-comparison}
\end{table*}

From Table~\ref{tab:dataset-benchmark-comparison}, we see that training on BioAlchemy with verifiable RL results in the best average performance (40.41\%). In addition, we find that RL is well suited for our dataset; given the same dataset split, training with RL on BioAlchemy results in an overall gain of +4.17\% compared to training with SFT. On SFT, BioAlchemy outperforms NaturalReasoning and performs as well as TextbookReasoning, despite distilling from a smaller teacher model (Llama-3.3 70B Instruct versus DeepSeek-V3). Similar to BioAlchemy, we also find that Nemotron-Science-v1 has strong performance when trained with RL, performing better than the other SFT datasets. We attribute this result to Nemotron-Science-v1 and BioAlchemy both having denser MeSH topics and stronger subtopic alignment, as seen in Table~\ref{tab:mesh-topic-coverage}. This supports our hypothesis that better subtopic alignment with the scientific distribution can increase reasoning performance. 

\subsection{Optimizing subtopic distributions}\label{sec:alpha-distribution-section}

We examine whether performance on BioAlchemy can be improved by reweighting its distribution to upsample reasoning problems at the tail categories, which can be seen as a form of data augmentation in the subtopic space. To do this, we trained on four distributional variants of the BioAlchemy dataset from Section~\ref{sec:exponential-smoothing}, $\{\mathcal{S}_{\alpha=0.3},\mathcal{S}_{\alpha=0.5},\mathcal{S}_{\alpha=0.7},\mathcal{S}_{\alpha=1.0} \}$. Decreasing the $\alpha$ results in a dataset $\mathcal{S}_{\alpha}$ that is more uniform in subtopics, but produces a training distribution that is less aligned with the biology subtopic distribution $P_{\text{Bio}}$.

\FloatBarrier

\begin{table}[!htbp]
\centering
\small
\setlength{\tabcolsep}{4pt}
\renewcommand{\arraystretch}{1.10}
\begin{threeparttable}
\resizebox{\linewidth}{!}{%
\begin{tabular}{lcccccccc}
\toprule
\textbf{Dataset}
& \makecell{\textbf{Alignment}}
& \makecell{\textbf{Upsampling}}
& \makecell{\textbf{ProtocolQA}}
& \makecell{\textbf{SeqQA}}
& \makecell{\textbf{Cloning}\\\textbf{Scenarios}}
& \makecell{\textbf{PubMedQA}}
& \makecell{\textbf{GPQA-Bio}}
& \makecell{\textbf{General Avg.}} \\
\midrule
\textbf{Baseline}
& N/A
& N/A
& 42.69$\pm$2.06\%
& 8.42$\pm$0.43\%
& 5.76$\pm$2.16\%
& \textbf{69.00$\pm$0.35\%}
& 42.63$\pm$1.19\%
& 33.70$\pm$0.65\% \\

\midrule
$\boldsymbol{\mathcal{S}_{\alpha=0.3}}$
& Weakest
& Strongest
& \textbf{46.30$\pm$1.25\%}
& 19.63$\pm$0.98\%
& \textbf{12.12$\pm$3.96\%}
& 67.75$\pm$0.28\%
& 53.16$\pm$2.78\%
& 39.79$\pm$1.02\% \\

$\boldsymbol{\mathcal{S}_{\alpha=0.5}}$
& Weak
& Strong
& 45.83$\pm$2.32\%
& 18.73$\pm$0.85\%
& 11.52$\pm$3.03\%
& 68.09$\pm$0.28\%
& 57.89$\pm$0.00\%
& 40.41$\pm$0.78\% \\

$\boldsymbol{\mathcal{S}_{\alpha=0.7}}$
& Strong
& Weak
& 44.63$\pm$1.97\%
& \textbf{20.13$\pm$0.75\%}
& 11.82$\pm$3.14\%
& 68.65$\pm$0.51\%
& 51.58$\pm$2.38\%
& 39.36$\pm$0.90\% \\

$\boldsymbol{\mathcal{S}_{\alpha=1.0}}$
& Strongest
& Weakest
& 45.93$\pm$1.60\%
& 19.98$\pm$0.73\%
& 10.91$\pm$3.57\%
& 68.83$\pm$0.30\%
& \textbf{63.16$\pm$0.00\%}
& \textbf{41.76$\pm$0.80\%} \\

\midrule
\textbf{Best}
& $\alpha=1.0$
& $\alpha=0.3$
& 46.30$\pm$1.25\%
& 20.13$\pm$0.75\%
& 12.12$\pm$3.96\%
& 69.00$\pm$0.35\%
& 63.16$\pm$0.00\%
& 41.76$\pm$0.80\% \\
\bottomrule
\end{tabular}%
}
\caption{Accuracy from training on datasets with stronger tail categories. Strongest alignment with $P_\text{Bio}$ leads to best performance. Baseline represents base Qwen3-8B results.}
\label{tab:alpha-rl-results}
\end{threeparttable}
\end{table}

\FloatBarrier

From Table~\ref{tab:alpha-rl-results}, we find that training on $\mathcal{S}_{\alpha=1.0}$ leads to the best overall performance, showing that strongly aligning with the subtopic distribution yields effective learning. The other datasets $\{\mathcal{S}_{\alpha=0.3}, \mathcal{S}_{\alpha=0.5}, \mathcal{S}_{\alpha=0.7}\}$ did not meaningfully differ in average performance due to strong overlap in confidence intervals. Therefore, we find that the majority of learning occurs in the initial calibration of the dataset itself. This is seen in $\text{Baseline} \rightarrow \mathcal{S}_{\alpha=1.0}$,  which has a gain of $+8.06\%$ in the general average. Further attempts at optimizing the distribution through upsampling the tail does not lead to better performance, and may even lead to degraded performance ($\mathcal{S}_{\alpha=1.0} \rightarrow \mathcal{S}_{\alpha=0.3}$ results in $-1.97\%$ in general average).  

\FloatBarrier

\subsection{Comparison with reasoning models}\label{sec:final-rl}

For our final experiment, we train BioAlchemist-8B and compare it to other reasoning models of similar size. We train for 1 epoch on the full dataset size (150K) for the largest topic dataset split, $\mathcal{S}_{\alpha=0.5}$. Although Section~\ref{sec:alpha-distribution-section} found that dataset split $\mathcal{S}_{\alpha=1.0}$ had the best performance at fixed dataset size from having the strongest subtopic alignment, we instead use $\mathcal{S}_{\alpha=0.5}$. We make this choice since $\mathcal{S}_{\alpha=0.5}$ is nearly $1.83\times$ as large as $\mathcal{S}_{\alpha=1.0}$, allowing us to assess whether scaling reasoning data can offset slightly worse performance from weaker alignment. This choice also lets us evaluate performance on the largest topic-balanced dataset variant available.

\FloatBarrier

\begin{table}[!htp]
\centering
\small
\setlength{\tabcolsep}{4pt}
\renewcommand{\arraystretch}{1.10}
\resizebox{\linewidth}{!}{%
\begin{tabular}{lcccccc}
\toprule
\textbf{Model}
& \makecell{\textbf{ProtocolQA}}
& \makecell{\textbf{SeqQA}}
& \makecell{\textbf{Cloning}\\\textbf{Scenarios}}
& \makecell{\textbf{PubMedQA}}
& \makecell{\textbf{GPQA-Bio}}
& \makecell{\textbf{Overall Avg.}} \\
\midrule
\textbf{BioAlchemist-8B 150K}
& 46.20$\pm$1.89\%
& \textbf{22.32$\pm$0.91\%}
& \textbf{15.15$\pm$4.57\%}
& 68.32$\pm$0.48\%
& \textbf{62.11$\pm$2.38\%}
& \textbf{42.82$\pm$1.12\%} \\

\midrule

\textbf{BioAlchemist-8B 50K}
& 45.83$\pm$2.32\%
& 18.73$\pm$0.85\%
& 11.52$\pm$3.03\%
& 68.09$\pm$0.28\%
& 57.89$\pm$0.00\%
& 40.41$\pm$0.78\% \\
\textbf{Qwen3-8B}
& 42.69$\pm$2.06\%
& 8.42$\pm$0.43\%
& 5.76$\pm$2.16\%
& 69.00$\pm$0.35\%
& 42.63$\pm$1.19\%
& 33.70$\pm$0.65\% \\
\textbf{DeepSeek-R1-Llama-8B}
& 33.61$\pm$2.58\%
& 4.97$\pm$0.41\%
& 10.91$\pm$2.09\%
& 26.89$\pm$0.58\%
& 5.26$\pm$0.00\%
& 16.33$\pm$0.68\% \\
\textbf{GPT-OSS-20B}
& \textbf{52.78$\pm$2.25\%}
& 18.12$\pm$0.79\%
& 7.27$\pm$1.83\%
& \textbf{72.79$\pm$0.24\%}
& 55.79$\pm$3.17\%
& 41.35$\pm$0.88\% \\
\midrule
\textbf{Best}
& 52.78$\pm$2.25\%
& 22.32$\pm$0.91\%
& 15.15$\pm$4.57\%
& 72.79$\pm$0.24\%
& 62.11$\pm$2.38\%
& 42.82$\pm$1.12\% \\
\bottomrule
\end{tabular}%
}
\caption{Final RL model performance across benchmarks. BioAlchemy-8B 150K performs best, and scales with more data.}
\label{tab:final-rl-results}
\end{table}

From Table~\ref{tab:final-rl-results}, we see that BioAlchemist-8B 150K outperforms other reasoning models of similar size. In particular, BioAlchemist-8B 150K performs best on SeqQA, Cloning Scenarios, and GPQA-Bio, indicating strong abilities in both graduate-level biology and applied sequencing and cloning tasks. Furthermore, we find that our model's performance improves across all benchmarks when trained on a larger amount of data, suggesting potential for improved performance through scaling data. Finally, we outperform GPT-OSS-20B \citep{openai2025gptoss} on the overall average, suggesting our model has potential to compete with larger reasoning models on biological reasoning.

One surprising outlier is DeepSeek-R1-Llama-8B, which performs noticeably worse on PubMedQA and GPQA-Bio given the same inference budget. One possible explanation is that this model requires significantly more test-time compute compared to other models to reach the same answer. Certain models such as DeepSeek-R1 and QwQ have been observed to exhibit overthinking on tasks such as mathematics \citep{chen2025overthinking}, which may also arise in comparatively less explored domains such as scientific biology. A more systematic evaluation of how test-time compute affects performance on biology benchmarks is left for future work.

\section{Conclusion}

We introduce BioAlchemy, a dataset and pipeline for sourcing reasoning problems from scientific papers and aligning them with scientific biology at the subtopic level. As part of constructing our dataset, we present a sampling method that better reflects the reasoning demands of modern biological research at the subtopic level. We find our dataset has higher MeSH topic density and coverage than the biology questions from other reasoning datasets. We demonstrate experimentally that training on BioAlchemy yields stronger performance gains than training on the biology subsets of other reasoning datasets. We also demonstrate how one can additionally optimize the BioAlchemy dataset through exponential smoothing strategies. Finally, we present BioAlchemist-8B, which has improved performance on applied scientific tasks in biology. This work lays the groundwork for next-generation biological reasoning models capable of assisting scientists working at the frontier of biological knowledge.

\subsubsection*{Acknowledgments}
Funding support from the US Department of Energy's Biological and Environmental Research (BER), Advanced Research Projects Agency for Health (ARPA-H) and Coalition for Epidemic Preparedness Innovations (CEPI). The compute resources for the project were provided by the U.S. Department of Energy’s (DOE) Innovative and Novel Computational Impact on Theory and Experiment (INCITE) Program. This research used resources from the Argonne Leadership Computing Facility, a U.S. DOE Office of Science user facility at Argonne National Laboratory, which is supported by the Office of Science of the U.S. DOE under Contract No. DE-AC02-06CH11357. We gratefully acknowledge the American Society for Microbiology (ASM) for granting access to their published literature in support of this work, and for their commitment to advancing biological research responsibly.

\clearpage

\bibliographystyle{colm2026_conference}
\bibliography{colm2026_conference}

@misc{mesh_biology,
  author = {{National Library of Medicine}},
  title  = {Biology},
  year   = {2026},
  url    = {https://meshb.nlm.nih.gov/record/ui?ui=D001695},
  note   = {MeSH descriptor, tree code H01.158.273.}
}

@article{chawla2025quantifying,
  title={Quantifying the importance of data alignment in downstream model performance},
  author={Chawla, Krrish and Sahai, Aryan and DePavia, Mario and Sundar, Sudharsan and Miranda, Brando and Obbad, Elyas and Koyejo, Sanmi},
  journal={arXiv preprint arXiv:2501.08496},
  year={2025}
}

@inproceedings{chen2025overthinking,
  title={Do NOT think that much for 2+ 3=? on the overthinking of long reasoning models},
  author={Chen, Xingyu and Xu, Jiahao and Liang, Tian and He, Zhiwei and Pang, Jianhui and Yu, Dian and Song, Linfeng and Liu, Qiuzhi and Zhou, Mengfei and Zhang, Zhuosheng and others},
  booktitle={Forty-second International Conference on Machine Learning},
  year={2025}
}

@inproceedings{devlin2019bert,
  title={Bert: Pre-training of deep bidirectional transformers for language understanding},
  author={Devlin, Jacob and Chang, Ming-Wei and Lee, Kenton and Toutanova, Kristina},
  booktitle={Proceedings of the 2019 conference of the North American chapter of the association for computational linguistics: human language technologies, volume 1 (long and short papers)},
  pages={4171--4186},
  year={2019}
}

@article{verl_github,
  title   = {HybridFlow: A Flexible and Efficient RLHF Framework},
  author  = {Guangming Sheng and Chi Zhang and Zilingfeng Ye and Xibin Wu and Wang Zhang and Ru Zhang and Yanghua Peng and Haibin Lin and Chuan Wu},
  year    = {2024},
  journal = {arXiv preprint arXiv: 2409.19256}
}

@article{blakeney2024does,
  title={Does your data spark joy? performance gains from domain upsampling at the end of training},
  author={Blakeney, Cody and Paul, Mansheej and Larsen, Brett W and Owen, Sean and Frankle, Jonathan},
  journal={arXiv preprint arXiv:2406.03476},
  year={2024}
}

@article{mizrahi2025language,
  title={Language models improve when pretraining data matches target tasks},
  author={Mizrahi, David and Larsen, Anders Boesen Lindbo and Allardice, Jesse and Petryk, Suzie and Gorokhov, Yuri and Li, Jeffrey and Fang, Alex and Gardner, Josh and Gunter, Tom and Dehghan, Afshin},
  journal={arXiv preprint arXiv:2507.12466},
  year={2025}
}

@misc{futurehouse_labbench_repo_2025,
  author = {{FutureHouse}},
  title  = {{LAB-Bench}: Measuring Capabilities of Language Models for Biology Research},
  year   = {2025},
  url    = {https://github.com/Future-House/LAB-Bench},
  note   = {GitHub repository.}
}

@article{openai2025gptoss,
  title={gpt-oss-120b \& gpt-oss-20b model card},
  author={Agarwal, Sandhini and Ahmad, Lama and Ai, Jason and Altman, Sam and Applebaum, Andy and Arbus, Edwin and others},
  journal={arXiv preprint arXiv:2508.10925},
  year={2025}
}

@inproceedings{jin2019pubmedqa,
  title={Pubmedqa: A dataset for biomedical research question answering},
  author={Jin, Qiao and Dhingra, Bhuwan and Liu, Zhengping and Cohen, William and Lu, Xinghua},
  booktitle={Proceedings of the 2019 conference on empirical methods in natural language processing and the 9th international joint conference on natural language processing (EMNLP-IJCNLP)},
  pages={2567--2577},
  year={2019}
}

@inproceedings{rein2024gpqa,
  title={Gpqa: A graduate-level google-proof q\&a benchmark},
  author={Rein, David and Hou, Betty Li and Stickland, Asa Cooper and Petty, Jackson and Pang, Richard Yuanzhe and Dirani, Julien and Michael, Julian and Bowman, Samuel R},
  booktitle={First conference on language modeling},
  year={2024}
}

@inproceedings{odonoghue2023bioplanner,
  title={BioPlanner: automatic evaluation of LLMs on protocol planning in biology},
  author={O’Donoghue, Odhran and Shtedritski, Aleksandar and Ginger, John and Abboud, Ralph and Ghareeb, Ali and Rodriques, Samuel},
  booktitle={Proceedings of the 2023 Conference on Empirical Methods in Natural Language Processing},
  pages={2676--2694},
  year={2023}
}

@article{liu2025understanding,
  title={Understanding r1-zero-like training: A critical perspective},
  author={Liu, Zichen and Chen, Changyu and Li, Wenjun and Qi, Penghui and Pang, Tianyu and Du, Chao and Lee, Wee Sun and Lin, Min},
  journal={arXiv preprint arXiv:2503.20783},
  year={2025}
}

@article{ouyang2022training,
  title={Training language models to follow instructions with human feedback},
  author={Ouyang, Long and Wu, Jeffrey and Jiang, Xu and Almeida, Diogo and Wainwright, Carroll and Mishkin, Pamela and Zhang, Chong and Agarwal, Sandhini and Slama, Katarina and Ray, Alex and others},
  journal={Advances in neural information processing systems},
  volume={35},
  pages={27730--27744},
  year={2022}
}

@article{liu2020multilingual,
  title={Multilingual denoising pre-training for neural machine translation},
  author={Liu, Yinhan and Gu, Jiatao and Goyal, Naman and Li, Xian and Edunov, Sergey and Ghazvininejad, Marjan and Lewis, Mike and Zettlemoyer, Luke},
  journal={Transactions of the Association for Computational Linguistics},
  volume={8},
  pages={726--742},
  year={2020},
  publisher={MIT Press One Rogers Street, Cambridge, MA 02142-1209, USA journals-info~…}
}

@inproceedings{conneau2020unsupervised,
  title={Unsupervised cross-lingual representation learning at scale},
  author={Conneau, Alexis and Khandelwal, Kartikay and Goyal, Naman and Chaudhary, Vishrav and Wenzek, Guillaume and Guzm{\'a}n, Francisco and Grave, Edouard and Ott, Myle and Zettlemoyer, Luke and Stoyanov, Veselin},
  booktitle={Proceedings of the 58th annual meeting of the association for computational linguistics},
  pages={8440--8451},
  year={2020}
}

@inproceedings{xue2021mt5,
  title={mT5: A massively multilingual pre-trained text-to-text transformer},
  author={Xue, Linting and Constant, Noah and Roberts, Adam and Kale, Mihir and Al-Rfou, Rami and Siddhant, Aditya and Barua, Aditya and Raffel, Colin},
  booktitle={Proceedings of the 2021 conference of the North American chapter of the association for computational linguistics: Human language technologies},
  pages={483--498},
  year={2021}
}

@article{zhu2025probing,
  title={Probing the critical point (critpt) of ai reasoning: a frontier physics research benchmark},
  author={Zhu, Minhui and Tian, Minyang and Yang, Xiaocheng and Zhou, Tianci and Yuan, Lifan and Zhu, Penghao and Chertkov, Eli and Liu, Shengyan and Du, Yufeng and Ji, Ziming and others},
  journal={arXiv preprint arXiv:2509.26574},
  year={2025}
}

@article{khattab2023dspy,
  title={Dspy: Compiling declarative language model calls into self-improving pipelines},
  author={Khattab, Omar and Singhvi, Arnav and Maheshwari, Paridhi and Zhang, Zhiyuan and Santhanam, Keshav and Vardhamanan, Sri and Haq, Saiful and Sharma, Ashutosh and Joshi, Thomas T and Moazam, Hanna and others},
  journal={arXiv preprint arXiv:2310.03714},
  year={2023}
}

@misc{plos_genetics_journal,
  author = {{Public Library of Science}},
  title  = {{PLoS Genetics}},
  year   = {2026},
  url    = {https://journals.plos.org/plosgenetics/},
}

@misc{plos_compbio_journal,
  author = {{Public Library of Science}},
  title  = {{PLoS Computational Biology}},
  year   = {2026},
  url    = {https://journals.plos.org/ploscompbiol/},
}

@article{papersearchqa,
  title={PaperSearchQA: Learning to Search and Reason over Scientific Papers with RLVR},
  author={Burgess, James and Hansen, Jan N and Peng, Duo and Zhang, Yuhui and Lozano, Alejandro and Sun, Min Woo and Lundberg, Emma and Yeung-Levy, Serena},
  journal={arXiv preprint arXiv:2601.18207},
  year={2026}
}

@article{openmathreasoning,
  title={Aimo-2 winning solution: Building state-of-the-art mathematical reasoning models with openmathreasoning dataset},
  author={Moshkov, Ivan and Hanley, Darragh and Sorokin, Ivan and Toshniwal, Shubham and Henkel, Christof and Schifferer, Benedikt and Du, Wei and Gitman, Igor},
  journal={arXiv preprint arXiv:2504.16891},
  year={2025}
}

@article{numina_math_datasets,
  title={Numinamath: The largest public dataset in ai4maths with 860k pairs of competition math problems and solutions},
  author={Li, Jia and Beeching, Edward and Tunstall, Lewis and Lipkin, Ben and Soletskyi, Roman and Huang, Shengyi and Rasul, Kashif and Yu, Longhui and Jiang, Albert Q and Shen, Ziju and others},
  journal={Hugging Face repository},
  volume={13},
  number={9},
  pages={9},
  year={2024}
}

@article{huang2025gemini,
  title={Gemini 2.5 pro capable of winning gold at imo 2025},
  author={Huang, Yichen and Yang, Lin F},
  journal={arXiv preprint arXiv:2507.15855},
  volume={7},
  pages={7},
  year={2025}
}

@article{adaparse,
  title={AdaParse: An Adaptive Parallel PDF Parsing and Resource Scaling Engine},
  author={Siebenschuh, Carlo and Hippe, Kyle and Gokdemir, Ozan and Brace, Alexander and Khan, Arham M and Hossain, Khalid and Babuji, Yadu and Chia, Nicholas and Vishwanath, Venkatram and Ramanathan, Arvind and others},
  journal={Proceedings of Machine Learning and Systems},
  volume={7},
  year={2025}
}

@misc{ASM,
  author = {{American Society for Microbiology}},
  title  = {American Society for Microbiology},
  year   = {2025},
  url    = {https://asm.org},
}

@article{semanticscholar,
  title={The semantic scholar open data platform},
  author={Kinney, Rodney and Anastasiades, Chloe and Authur, Russell and Beltagy, Iz and Bragg, Jonathan and Buraczynski, Alexandra and Cachola, Isabel and Candra, Stefan and Chandrasekhar, Yoganand and Cohan, Arman and others},
  journal={arXiv preprint arXiv:2301.10140},
  year={2023}
}

@article{naturalreasoning,
  title={Naturalreasoning: Reasoning in the wild with 2.8 m challenging questions, 2025},
  author={Yuan, Weizhe and Yu, Jane and Jiang, Song and Padthe, Karthik and Li, Yang and Wang, Dong and Kulikov, Ilia and Cho, Kyunghyun and Tian, Yuandong and Weston, Jason E and others},
  journal={URL https://arxiv. org/abs/2502.13124}, 
  year={2025}
}

@article{megascience,
  title={Megascience: Pushing the frontiers of post-training datasets for science reasoning},
  author={Fan, Run-Ze and Wang, Zengzhi and Liu, Pengfei},
  journal={arXiv preprint arXiv:2507.16812},
  year={2025}
}

@article{yang2025qwen3,
  title={Qwen3 technical report},
  author={Yang, An and Li, Anfeng and Yang, Baosong and Zhang, Beichen and Hui, Binyuan and Zheng, Bo and Yu, Bowen and Gao, Chang and Huang, Chengen and Lv, Chenxu and others},
  journal={arXiv preprint arXiv:2505.09388},
  year={2025}
}

@article{wen2025reinforcementlearningverifiablerewards,
  title={Reinforcement learning with verifiable rewards implicitly incentivizes correct reasoning in base llms},
  author={Wen, Xumeng and Liu, Zihan and Zheng, Shun and Ye, Shengyu and Wu, Zhirong and Wang, Yang and Xu, Zhijian and Liang, Xiao and Li, Junjie and Miao, Ziming and others},
  journal={arXiv preprint arXiv:2506.14245},
  year={2025}
}

@article{sayers2009eutilities,
  title={A General Introduction to the E-utilities},
  author={Sayers, Eric},
  journal={Entrez Programming Utilities Help [Internet]. Bethesda (MD): National Center for Biotechnology Information (US)},
  year={2010}
}

@misc{pubmed,
  title   = {{PubMed}},
  author  = {{National Library of Medicine}},
  year    = {2026},
  url     = {https://pubmed.ncbi.nlm.nih.gov/}
}

@misc{nemotron_science_v1,
  author = {{NVIDIA}},
  title  = {{Nemotron-Science-v1}},
  year   = {2025},
  url    = {https://huggingface.co/datasets/nvidia/Nemotron-Science-v1}
}

@article{deepseek,
  title={DeepSeek-R1 incentivizes reasoning in LLMs through reinforcement learning},
  author={Guo, Daya and Yang, Dejian and Zhang, Haowei and Song, Junxiao and Wang, Peiyi and Zhu, Qihao and Xu, Runxin and Zhang, Ruoyu and Ma, Shirong and Bi, Xiao and others},
  journal={Nature},
  volume={645},
  number={8081},
  pages={633--638},
  year={2025},
  publisher={Nature Publishing Group UK London}
}

@article{labbench,
  title={Lab-bench: Measuring capabilities of language models for biology research},
  author={Laurent, Jon M and Janizek, Joseph D and Ruzo, Michael and Hinks, Michaela M and Hammerling, Michael J and Narayanan, Siddharth and Ponnapati, Manvitha and White, Andrew D and Rodriques, Samuel G},
  journal={arXiv preprint arXiv:2407.10362},
  year={2024}
}

@article{fallahpour2025bioreason,
  title={Bioreason: Incentivizing multimodal biological reasoning within a dna-llm model},
  author={Fallahpour, Adibvafa and Magnuson, Andrew and Gupta, Purav and Ma, Shihao and Naimer, Jack and Shah, Arnav and Duan, Haonan and Ibrahim, Omar and Goodarzi, Hani and Maddison, Chris J and others},
  journal={arXiv preprint arXiv:2505.23579},
  year={2025}
}

@article{liu2025bioprobench,
  title={Bioprobench: Comprehensive dataset and benchmark in biological protocol understanding and reasoning},
  author={Liu, Yuyang and Lv, Liuzhenghao and Zhang, Xiancheng and Yuan, Jingya Wang Li and Tian, Yonghong},
  journal={arXiv preprint arXiv:2505.07889},
  year={2025}
}

@inproceedings{zhang-etal-2025-capacitygap,
  title={Towards the law of capacity gap in distilling language models},
  author={Zhang, Chen and Li, Qiuchi and Song, Dawei and Ye, Zheyu and Gao, Yan and Hu, Yao},
  booktitle={Proceedings of the 63rd Annual Meeting of the Association for Computational Linguistics (Volume 1: Long Papers)},
  pages={22504--22528},
  year={2025}
}

@article{dapo,
  title={Dapo: An open-source llm reinforcement learning system at scale},
  author={Yu, Qiying and Zhang, Zheng and Zhu, Ruofei and Yuan, Yufeng and Zuo, Xiaochen and Yue, Yu and Dai, Weinan and Fan, Tiantian and Liu, Gaohong and Liu, Lingjun and others},
  journal={arXiv preprint arXiv:2503.14476},
  year={2025}
}

@article{shao2024deepseekmath,
  title={Deepseekmath: Pushing the limits of mathematical reasoning in open language models},
  author={Shao, Zhihong and Wang, Peiyi and Zhu, Qihao and Xu, Runxin and Song, Junxiao and Bi, Xiao and Zhang, Haowei and Zhang, Mingchuan and Li, YK and Wu, Yang and others},
  journal={arXiv preprint arXiv:2402.03300},
  year={2024}
}

@inproceedings{HiPerRag,
  title={HiPerRAG: High-performance retrieval augmented generation for scientific insights},
  author={Gokdemir, Ozan and Siebenschuh, Carlo and Brace, Alexander and Wells, Azton and Hsu, Brian and Hippe, Kyle and Setty, Priyanka and Ajith, Aswathy and Pauloski, J Gregory and Sastry, Varuni and others},
  booktitle={Proceedings of the Platform for Advanced Scientific Computing Conference},
  pages={1--13},
  year={2025}
}

@inproceedings{gururangan-etal-2020-dont,
  title={Don’t stop pretraining: Adapt language models to domains and tasks},
  author={Gururangan, Suchin and Marasovi{\'c}, Ana and Swayamdipta, Swabha and Lo, Kyle and Beltagy, Iz and Downey, Doug and Smith, Noah A},
  booktitle={Proceedings of the 58th annual meeting of the association for computational linguistics},
  pages={8342--8360},
  year={2020}
}

@inproceedings{AutomatedMCQA,
  title={Automated MCQA Benchmarking at Scale: Evaluating Reasoning Traces as Retrieval Sources for Domain Adaptation of Small Language Models},
  author={Gokdemir, Ozan and Getty, Neil and Underwood, Robert and Madireddy, Sandeep and Cappello, Franck and Ramanathan, Arvind and Foster, Ian T and Stevens, Rick L},
  booktitle={Proceedings of the SC'25 Workshops of the International Conference for High Performance Computing, Networking, Storage and Analysis},
  pages={545--552},
  year={2025}
}

\clearpage

\appendix
\section{Appendix}

\subsection{GRPO settings}

\begin{table}[!htp]
\centering
\small
\setlength{\tabcolsep}{6pt}
\renewcommand{\arraystretch}{1.30}

\begin{threeparttable}
\resizebox{\linewidth}{!}{%
\begin{tabular}{lccccccc}
\toprule
\textbf{Training} 
& \textbf{Rollouts} 
& \textbf{Max Response Len} 
& \textbf{Temperature}
& \textbf{Clip Ratio Range}
& \textbf{Normalize Adv.}
& \textbf{Entropy Coef.}
& \textbf{Loss Aggr.}
\\
\midrule
GRPO
& 5
& 4096
& 0.6
& [0.2, 0.28]
& False
& 0.0005
& Token-mean
\\
\bottomrule
\end{tabular}%
}

\caption{GRPO-specific training configurations used across all experiments. Settings are applied using the VeRL framework.}
\label{tab:training-config-dataset-compare}
\end{threeparttable}
\end{table}

\subsection{Reasoning dataset comparison}\label{sec:appendix-dataset-comparison}

Due to lack of access to DeepSeek-V3, we leverage Llama-3.3 70B Instruct to provide reasoning traces for our SFT experiments. We sampled 50K questions from each dataset. This is to best match the scale of the other SFT datasets, with NaturalReasoning and TextbookReasoning each having around 50K biology questions, and Nemotron-Science-v1 having around 29K biology questions. For GRPO, we use a max response length of 4096 and max input prompt length of 1024 for training, and use an equivalent of 5120 max tokens for SFT to fix aggregate tokens used during training. 

\begin{table}[!htp]
\centering
\small
\setlength{\tabcolsep}{6pt}
\renewcommand{\arraystretch}{1.30}

\begin{threeparttable}
\resizebox{\linewidth}{!}{%
\begin{tabular}{lccccccccc}
\toprule
\textbf{Dataset} 
& \textbf{LR} 
& \textbf{LR-Scheduler} 
& \textbf{Batch Size}
& \textbf{Epochs}
& \textbf{Max Tokens}
& \textbf{Loss}
& \textbf{Dataset Size}
& \textbf{Teacher Model}
\\
\midrule
BioAlchemy RL
& 4e-6
& cosine
& 128
& 1
& 5120
& GRPO
& 50K
& N/A
\\
BioAlchemy SFT
& 4e-6
& cosine
& 128
& 1
& 5120
& SFT
& 50K
& Llama-3.3 70B Instruct
\\
Nemotron-Science-v1
& 4e-6
& cosine
& 128
& 1
& 5120
& GRPO
& 29K
& N/A
\\
NaturalReasoning
& 4e-6
& cosine
& 128
& 1
& 5120
& SFT
& 50K
& DeepSeek-R1
\\
TextbookReasoning
& 4e-6
& cosine
& 128
& 1
& 5120
& SFT
& 50K
& DeepSeek-V3
\\
\bottomrule
\end{tabular}%
}

\caption{Table containing training configurations for GRPO and SFT experiments.}
\label{tab:training-config-dataset-compare-second-with-same-label}
\end{threeparttable}
\end{table}

\subsection{Distribution optimization and RL}\label{sec:appendix-distribution-comparison}

$\mathcal{S}^{\text{Final}}_{\alpha=0.5}$ refers to the training run used in Section~\ref{sec:final-rl}, while $\mathcal{S}_{\alpha}$ refers to training distributions used in Section~\ref{sec:alpha-distribution-section}.

\begin{table}[!htp]
\centering
\small
\setlength{\tabcolsep}{6pt}
\renewcommand{\arraystretch}{1.30}

\begin{threeparttable}
\resizebox{\linewidth}{!}{%
\begin{tabular}{lccccccccc}
\toprule
\textbf{Dataset} 
& \textbf{LR} 
& \textbf{LR-Scheduler} 
& \textbf{Batch Size}
& \textbf{Epochs}
& \textbf{Max Tokens}
& \textbf{Max Response Len}
& \textbf{Loss}
& \textbf{Dataset Size}
\\
\midrule
\textbf{$\mathcal{S}_{\alpha=0.3}$}
& 4e-6
& cosine
& 128
& 1
& 5120
& 4096
& GRPO
& 60K
\\
\textbf{$\mathcal{S}_{\alpha=0.5}$}
& 4e-6
& cosine
& 128
& 1
& 5120
& 4096
& GRPO
& 50K
\\
\textbf{$\mathcal{S}_{\alpha=0.7}$}
& 4e-6
& cosine
& 128
& 1
& 5120
& 4096
& GRPO
& 60K
\\
\textbf{$\mathcal{S}_{\alpha=1.0}$}
& 4e-6
& cosine
& 128
& 1
& 5120
& 4096
& GRPO
& 60K
\\
$\mathcal{S}^{\text{Final}}_{\alpha=0.5}$
& 4e-6
& cosine
& 256
& 1
& 5120
& 4096
& GRPO
& 150K
\\
\bottomrule
\end{tabular}%
}

\caption{Training configurations for distribution optimization experiments and final RL experiment.}
\label{tab:training-config-distribution-opt}
\end{threeparttable}
\end{table}

\clearpage

\subsection{Inference settings}

Due to randomness in vLLM from data parallelism, we compute confidence intervals over 10 runs per benchmark for a better estimate on accuracy. 

\begin{table}[!htbp]
\centering
\small
\setlength{\tabcolsep}{6pt}
\renewcommand{\arraystretch}{1.30}

\begin{threeparttable}
\resizebox{\linewidth}{!}{%
\begin{tabular}{lcccccc}
\toprule
\textbf{Benchmark} 
& \textbf{Max Response Len} 
& \textbf{Seed} 
& \textbf{Temperature} 
& \textbf{Top-p}
& \textbf{Metric}
& \textbf{N-runs}
\\
\midrule
ProtocolQA
& 4096
& 42
& 0.0
& 1.0
& EM
& 10
\\
SeqQA
& 4096
& 42
& 0.0
& 1.0
& EM
& 10
\\
Cloning Scenarios
& 4096
& 42
& 0.0
& 1.0
& EM
& 10
\\
PubMedQA
& 4096
& 42
& 0.0
& 1.0
& EM
& 10
\\
GPQA-Bio
& 4096
& 42
& 0.0
& 1.0
& EM
& 10
\\
\bottomrule
\end{tabular}%
}

\caption{Inference settings for benchmarking. Models were served via vLLM, using tensor parallelism and data parallelism for throughput.}
\label{tab:inference-settings}
\end{threeparttable}
\end{table}

\FloatBarrier

\subsection{Full topic distributions based on alpha values}

\FloatBarrier
\begin{table}[!htp]
\centering
\small
\resizebox{\textwidth}{!}{%
\begin{tabular}{c>{\raggedright\arraybackslash}lrrrrrr}
\toprule
& & \multicolumn{2}{c}{PubMed Biology} & \multicolumn{4}{c}{Smoothed \% (Upsample$\times$)} \\
\cmidrule(lr){3-4} \cmidrule(lr){5-8}
Rank & MeSH Biology Subcategory & Count & \% & $\alpha=1.0$ & $\alpha=0.7$ & $\alpha=0.5$ & $\alpha=0.3$ \\
\midrule
1 & Computational Biology & 95,179 & 45.32\% & 45.32\% (1.0$\times$) & 36.05\% (0.8$\times$) & 26.97\% (0.6$\times$) & 16.51\% (0.4$\times$) \\
2 & Genetics & 75,973 & 36.18\% & 36.18\% (1.0$\times$) & 30.79\% (0.9$\times$) & 24.10\% (0.7$\times$) & 15.43\% (0.4$\times$) \\
3 & Microbiology & 22,900 & 10.90\% & 10.90\% (1.0$\times$) & 13.30\% (1.2$\times$) & 13.23\% (1.2$\times$) & 10.77\% (1.0$\times$) \\
4 & Ecology & 5,385 & 2.56\% & 2.56\% (1.0$\times$) & 4.83\% (1.9$\times$) & 6.42\% (2.5$\times$) & 6.98\% (2.7$\times$) \\
5 & Molecular Biology & 3,944 & 1.88\% & 1.88\% (1.0$\times$) & 3.88\% (2.1$\times$) & 5.49\% (2.9$\times$) & 6.35\% (3.4$\times$) \\
6 & Synthetic Biology & 2,034 & 0.97\% & 0.97\% (1.0$\times$) & 2.44\% (2.5$\times$) & 3.94\% (4.1$\times$) & 5.21\% (5.4$\times$) \\
7 & Botany & 1,273 & 0.61\% & 0.61\% (1.0$\times$) & 1.76\% (2.9$\times$) & 3.12\% (5.1$\times$) & 4.53\% (7.5$\times$) \\
8 & Developmental Biology & 603 & 0.29\% & 0.29\% (1.0$\times$) & 1.04\% (3.6$\times$) & 2.15\% (7.5$\times$) & 3.62\% (12.6$\times$) \\
9 & Zoology & 547 & 0.26\% & 0.26\% (1.0$\times$) & 0.97\% (3.7$\times$) & 2.04\% (7.9$\times$) & 3.51\% (13.5$\times$) \\
10 & Neurobiology & 427 & 0.20\% & 0.20\% (1.0$\times$) & 0.82\% (4.0$\times$) & 1.81\% (8.9$\times$) & 3.26\% (16.0$\times$) \\
11 & Exobiology & 384 & 0.18\% & 0.18\% (1.0$\times$) & 0.76\% (4.2$\times$) & 1.71\% (9.4$\times$) & 3.16\% (17.3$\times$) \\
12 & Radiobiology & 324 & 0.15\% & 0.15\% (1.0$\times$) & 0.68\% (4.4$\times$) & 1.57\% (10.2$\times$) & 3.00\% (19.5$\times$) \\
13 & Parasitology & 292 & 0.14\% & 0.14\% (1.0$\times$) & 0.63\% (4.5$\times$) & 1.49\% (10.7$\times$) & 2.91\% (20.9$\times$) \\
14 & Cell Biology & 256 & 0.12\% & 0.12\% (1.0$\times$) & 0.57\% (4.7$\times$) & 1.40\% (11.5$\times$) & 2.80\% (22.9$\times$) \\
15 & Marine Biology & 204 & 0.10\% & 0.10\% (1.0$\times$) & 0.49\% (5.0$\times$) & 1.25\% (12.9$\times$) & 2.61\% (26.9$\times$) \\
16 & Laboratory Animal Science & 112 & 0.05\% & 0.05\% (1.0$\times$) & 0.32\% (6.0$\times$) & 0.93\% (17.4$\times$) & 2.18\% (40.9$\times$) \\
17 & Natural History & 87 & 0.04\% & 0.04\% (1.0$\times$) & 0.27\% (6.5$\times$) & 0.82\% (19.7$\times$) & 2.02\% (48.8$\times$) \\
18 & Photobiology & 47 & 0.02\% & 0.02\% (1.0$\times$) & 0.17\% (7.8$\times$) & 0.60\% (26.8$\times$) & 1.68\% (75.2$\times$) \\
19 & Cryobiology & 15 & 0.01\% & 0.01\% (1.0$\times$) & 0.08\% (11.0$\times$) & 0.34\% (47.4$\times$) & 1.19\% (167.2$\times$) \\
20 & Cytology & 13 & 0.01\% & 0.01\% (1.0$\times$) & 0.07\% (11.5$\times$) & 0.32\% (50.9$\times$) & 1.14\% (184.8$\times$) \\
21 & Sociobiology & 12 & 0.01\% & 0.01\% (1.0$\times$) & 0.07\% (11.8$\times$) & 0.30\% (53.0$\times$) & 1.12\% (195.5$\times$) \\
\midrule
& \textbf{Max/Min Ratio} & & 7,932$\times$ & 7,932$\times$ & 536$\times$ & 89$\times$ & 15$\times$ \\
& \textbf{Avg. Upsample (5 rarest)} & & 1.0$\times$ & 1.0$\times$ & 9.7$\times$ & 39.6$\times$ & 134.3$\times$ \\
\bottomrule
\end{tabular}%
}
\caption{Effect of exponential smoothing ($p^{\alpha}_{c_{j}} \propto n_{c_{j}}^{\alpha}$) on the PubMed Biology subcategory sampling distribution, where $c_{j}$ for $j\in[21]$ represents the 21 MeSH Biology subcategories. We adopt $\alpha = 0.5$ as a moderate compromise between the values used in multilingual language model pretraining: $\alpha = 0.7$ \citep{devlin2019bert, liu2020multilingual} and $\alpha = 0.3$ \citep{conneau2020unsupervised, xue2021mt5}. Note that $\alpha = 1.0$ recovers the original unsmoothed distribution. Total count for PubMed Biology is 210,011 articles. Upsample factor for each category is $\nicefrac{p_{c_j}^{\alpha}}{p_{c_j}^{\mathrm{orig}}}$, showing stronger upsampling as $\alpha$ decreases.}
\label{tab:smoothed_sampling_full}
\end{table}

\FloatBarrier

\clearpage

\subsection{Full topic distributions of biological reasoning datasets}

\FloatBarrier
\begin{table}[!htbp]
\centering
\resizebox{\textwidth}{!}{%
\begin{tabular}{clrrrrr}
\toprule
Rank & MeSH Biology Subcategory
& \makecell{PubMed Biology}
& \makecell{Textbook\\Reasoning}
& \makecell{Natural\\Reasoning}
& \makecell{Nemotron\\Science-v1}
& \makecell{BioAlchemy} \\
\midrule
1 & Computational Biology 
& 45.32\% 
& 0.57\% 
& 12.09\% 
& 7.44\% 
& 24.95\% \\
2 & Genetics 
& 36.18\% 
& 11.37\% 
& 18.30\% 
& 34.67\% 
& 36.89\% \\
3 & Microbiology 
& 10.90\% 
& 14.39\% 
& 7.19\% 
& 3.78\% 
& 16.75\% \\
4 & Ecology 
& 2.56\% 
& 2.51\% 
& 9.24\% 
& 0.99\% 
& 2.80\% \\
5 & Molecular Biology 
& 1.88\% 
& 12.59\% 
& 8.45\% 
& 40.41\% 
& 6.63\% \\
6 & Synthetic Biology 
& 0.97\% 
& 0.04\% 
& 0.63\% 
& 1.76\% 
& 0.50\% \\
7 & Botany 
& 0.61\% 
& 15.72\% 
& 2.86\% 
& 1.13\% 
& 0.79\% \\
8 & Developmental Biology 
& 0.29\% 
& 5.36\% 
& 3.84\% 
& 2.94\% 
& 2.71\% \\
9 & Zoology 
& 0.26\% 
& 11.86\% 
& 4.43\% 
& 0.67\% 
& 0.31\% \\
10 & Neurobiology 
& 0.20\% 
& 6.22\% 
& 11.77\% 
& 0.81\% 
& 4.57\% \\
11 & Exobiology 
& 0.18\% 
& 0.10\% 
& 5.72\% 
& 0.02\% 
& 0.06\% \\
12 & Radiobiology 
& 0.15\% 
& 1.12\% 
& 3.10\% 
& 0.11\% 
& 0.74\% \\
13 & Parasitology 
& 0.14\% 
& 4.53\% 
& 0.51\% 
& 0.10\% 
& 0.64\% \\
14 & Cell Biology 
& 0.12\% 
& 7.81\% 
& 4.86\% 
& 4.49\% 
& 1.32\% \\
15 & Marine Biology 
& 0.10\% 
& 1.08\% 
& 0.78\% 
& 0.15\% 
& 0.11\% \\
16 & Laboratory Animal Science 
& 0.05\% 
& 0.39\% 
& 0.21\% 
& 0.05\% 
& 0.09\% \\
17 & Natural History 
& 0.04\% 
& 0.66\% 
& 0.51\% 
& 0.01\% 
& 0.01\% \\
18 & Photobiology 
& 0.02\% 
& 0.74\% 
& 3.54\% 
& 0.39\% 
& 0.05\% \\
19 & Cryobiology 
& 0.01\% 
& 0.31\% 
& 0.56\% 
& 0.04\% 
& 0.01\% \\
20 & Cytology 
& 0.01\% 
& 2.39\% 
& 0.13\% 
& 0.02\% 
& 0.02\% \\
21 & Sociobiology 
& 0.01\% 
& 0.24\% 
& 1.30\% 
& 0.01\% 
& 0.06\% \\
\midrule
& \textbf{Total Samples} & \textbf{210,011} & \textbf{52,850} & \textbf{51,150} & \textbf{29,047} & \textbf{126,721} \\
& \textbf{TVD from PubMed} & \textbf{0.00} & \textbf{0.71} & \textbf{0.55} & \textbf{0.48} & \textbf{0.21} \\
\bottomrule
\end{tabular}
}%
\caption{MeSH Biology subcategory distributions across biological reasoning datasets
compared to PubMed Biology (2020--2024). TVD = total variation distance. This BioAlchemy dataset version uses the strongest alignment method (exponential smoothing of $\alpha=1.0$ detailed in Section~\ref{sec:exponential-smoothing}), yielding the lowest TVD with PubMed Biology.}
\label{tab:topic-distributions-full}
\end{table}
\FloatBarrier

\clearpage

\subsection{Prompt classifier}

Prompt optimization with \text{BootstrapFewShotWithRandomSearch}  \citep{khattab2023dspy} evaluated on 11 candidate prompt configurations. During optimization, the best-performing configuration achieved 61.6\% F1. On the validation set, the optimized classifier attained 72.7\% micro-averaged F1. This represents a substantial improvement over the baseline zero-shot configuration (37.0\% F1), corresponding to a relative improvement of 96\%. The transition from single-label to multi-label ground truth yielded a 23.4\% improvement in validation F1, indicating that a substantial portion of apparent classification errors under single-label evaluation represented correct identification of secondary category assignments.

\FloatBarrier
\begin{table}[!htbp]
\centering
\resizebox{0.7\linewidth}{!}{%
\begin{tabular}{lcc}
\toprule
\textbf{Configuration} & \textbf{Optimization F1} & \textbf{Validation F1} \\
\midrule
Zero-shot baseline     & N/A     & 37.0\% \\
Single-label training  & 52.7\% & 49.3\% \\
Multi-label training   & 61.6\% & 72.7\% \\
\bottomrule
\end{tabular}%
}
\caption{MeSH Biology classification performance before and after prompt optimization.}
\label{tab:mesh_results}
\end{table}

\FloatBarrier
\begin{table}[!htbp]
\centering
\resizebox{0.4\linewidth}{!}{%
\begin{tabular}{lr}
\toprule
\textbf{Metric} & \textbf{Value} \\
\midrule
Unique articles         & 1,474 \\
Total category labels   & 1,579 \\
Mean labels per article & 1.07 \\
Number of categories    & 21 \\
Training examples       & 714 \\
Validation examples     & 180 \\
\bottomrule
\end{tabular}%
}
\caption{Training data statistics for MeSH Biology classification.}
\label{tab:mesh_training_data}
\end{table}

\clearpage

\subsection{Reasoning question topic classification prompts}\label{sec:classification-prompts}

\begin{figure}[!htbp]
\centering
\begin{tcolorbox}[
  enhanced,
  colback=blue!10,
  colframe=blue!40,
  boxrule=0.6pt,
  arc=6pt,
  left=6pt,
  right=6pt,
  top=4pt,
  bottom=4pt,
  width=0.95\linewidth,
  title={\small\textbf{User}},fonttitle=\bfseries
]
\begin{lstlisting}[basicstyle=\small\rmfamily,breaklines=true,breakatwhitespace=true,columns=fullflexible]
You are a scientific domain classifier. Given a question-answer pair, 
determine whether it is primarily about BIOLOGY or not.

Biology includes: molecular biology, genetics, cell biology, ecology, microbiology, 
neurobiology, zoology, botany, developmental biology, evolutionary biology, 
biochemistry (when focused on biological systems), bioinformatics, marine biology, 
parasitology, immunology, physiology, anatomy, pharmacology (biological mechanisms), 
and related life sciences.

NOT biology: pure chemistry, physics, mathematics, computer science (unless
bioinformatics), engineering, geology (unless paleobiology), astronomy, pure statistics, 
materials science.

Question: {question}

Answer: {answer}

Respond with ONLY one word: "biology" or "not_biology"
\end{lstlisting}
\end{tcolorbox}
\caption{Prompt template for biology domain classification}
\label{fig:biology-domain-classification-prompt}
\end{figure}

\clearpage

\begin{figure}[!htbp]
\centering
\begin{tcolorbox}[
  enhanced,
  colback=blue!10,
  colframe=blue!40,
  boxrule=0.6pt,
  arc=6pt,
  left=6pt,
  right=6pt,
  top=4pt,
  bottom=4pt,
  width=0.95\linewidth,
  title={\small\textbf{User}},fonttitle=\bfseries
]
\begin{lstlisting}[basicstyle=\small\rmfamily,breaklines=true,breakatwhitespace=true,columns=fullflexible]
Classify a scientific abstract into MeSH Biology categories.

---

Follow the following format.

Abstract: Scientific abstract from a biology research paper
Reasoning: Let's think step by step in order to ${reasoning}
Categories: JSON array of MeSH category names that apply, e.g. ["Genetics", "Molecular 
Biology"]

---

Abstract: Because early-life stress is common and constitutes a strong risk factor for cognitive and mental health disorders, it has been the focus of a multitude of studies in humans and experimental models. Yet, we have an incomplete understanding of what is perceived as stressful by the developing brain, what aspects of stress influence brain maturation, what developmental ages are particularly vulnerable to stress, which molecules mediate the effects of stress on brain operations, and how transient stressful experiences can lead to enduring emotional and cognitive dysfunctions. Here, we discuss these themes, highlight the challenges and progress in resolving them, and propose new concepts and avenues for future research.
Reasoning: Let's think step by step in order to The abstract discusses the effects of early-life stress on brain development, cognitive function, and mental health, referencing both human and experimental model studies. It addresses how stress is perceived by the developing brain, the molecular mediators involved, and the long-term consequences of early stress on emotional and cognitive outcomes. These topics are central to the fields of neurobiology (specifically developmental neurobiology), behavioral science, and psychology, as they pertain to brain maturation, stress response, and mental health disorders.
Categories: ["Neurobiology", "Developmental Biology", "Behavioral Sciences", "Psychology"]

---

Abstract: Hormesis drives biological modifications from cells to higher levels of biological organization and emerges as a general basic principle of biology, integrating evolution, ecology, medicine, physiology, toxicology, and public health.
Reasoning: Let's think step by step in order to The abstract discusses hormesis as a fundamental biological principle that influences biological modifications at multiple levels of organization, from cells to higher-order systems. It also highlights the integrative role of hormesis across several biological and biomedical disciplines, including evolution, ecology, medicine, physiology, toxicology, and public health. This indicates that the topic is broad and interdisciplinary, touching on general biological processes, evolutionary biology, ecology, physiology, toxicology, and public health.
Categories: ["Biology", "Evolutionary Biology", "Ecology", "Physiology", "Toxicology", "Public Health"]
\end{lstlisting}
\end{tcolorbox}
\caption{Few-shot examples for MeSH Biology classification (part 1)}
\label{fig:mesh-classification-prompt-part1}
\end{figure}

\clearpage

\begin{figure}[!htbp]
\centering
\begin{tcolorbox}[
  enhanced,
  colback=blue!10,
  colframe=blue!40,
  boxrule=0.6pt,
  arc=6pt,
  left=6pt,
  right=6pt,
  top=4pt,
  bottom=4pt,
  width=0.95\linewidth,
  title={\small\textbf{User (cont.)}},fonttitle=\bfseries
]
\begin{lstlisting}[basicstyle=\small\rmfamily,breaklines=true,breakatwhitespace=true,columns=fullflexible]
Abstract: Permafrost represents 26% of terrestrial soil ecosystems; yet its biology, essentially microbiology, remains relatively unexplored. The permafrost environment is considered extreme because indigenous microorganisms must survive prolonged exposure to subzero temperatures and background radiation for geological time scales in a habitat with low water activity and extremely low rates of nutrient and metabolite transfer. Yet considerable numbers and biodiversity of bacteria exist in permafrost, some of which may be among the most ancient viable life on Earth. This review describes the permafrost environment as a microbial habitat and reviews recent studies examining microbial biodiversity found in permafrost as well as microbial growth and activity at ambient in situ subzero temperatures. These investigations suggest that functional microbial ecosystems exist within the permafrost environment and may have important implications on global biogeochemical processes as well as the search for past or extant life in permafrost presumably present on Mars and other bodies in our solar system.
Categories: ["Exobiology"]

---

Abstract: Submarines and spacecraft share several features that may promote the presence of fungi, including recirculated ventilation systems, moist areas, and close-quarters living. In this article, we introduce the idea of "submarine mycology" and explore how research on submarine fungi can inform the emerging field of astromycology. We highlight parallels in the fungal species present in both environments, while also noting key differences such as radiation exposure and microgravity. Arguing that submarines offer valuable lessons for spaceflight, we advocate for renewed research using modern genetic tools to characterize submarine fungi.
Categories: ["Exobiology"]

---

Abstract: Artemisia vulgaris L. belongs to the family Compositae, sub-family Corymbiferae. The genus Artemisia groups together almost 200 species, the most of them are native of Eurasia and Northern America steppe regions. Artemisia are wind-pollinated plants, the flowers do not secrete any nectar and are not visited by bees. Two species of Artemisia are widely spread through the Lyon region and are with Ambrosiaceae to blame for the pollinosis in summer and autumn: A. annua and A. vulgaris.
Categories: ["Botany"]

---

Abstract: Some basic concepts for the creation of the Swiss National Park were derived from observations made in Sri Lanka, Indonesia and New Caledonia. European researchers feared that the study of "virgin nature" would no longer be possible, as various species would soon become extinct under the combined influences of colonial practices and profit-oriented capitalism. While the motives for protecting nature originated from experiences made in the southern hemisphere, their scientific concept of conservation was based on European natural history and the related theories of evolution. In the light of this approach, endangered zoological and botanical species as well as "primitive" varieties of man were appreciated as "documents" to be preserved within their original environment for future scientific reference and research. Museum collections and reservations (parks) were two types of repositories connected to each other by the same objective.
Categories: ["Natural History"]
\end{lstlisting}
\end{tcolorbox}
\caption{Few-shot examples for MeSH Biology classification (part 2)}
\label{fig:mesh-classification-prompt-part2}
\end{figure}

\clearpage

\begin{figure}[!htbp]
\centering
\begin{tcolorbox}[
  enhanced,
  colback=blue!10,
  colframe=blue!40,
  boxrule=0.6pt,
  arc=6pt,
  left=6pt,
  right=6pt,
  top=4pt,
  bottom=4pt,
  width=0.95\linewidth,
  title={\small\textbf{User (cont.)}},fonttitle=\bfseries
]
\begin{lstlisting}[basicstyle=\small\rmfamily,breaklines=true,breakatwhitespace=true,columns=fullflexible]
Abstract: In the past 2 decades, structural biology has transformed from a single technique used on single proteins to a multimodal integrative approach. Recently, protein structure prediction algorithms have opened new avenues to address challenging biological questions.
Categories: ["Molecular Biology", "Computational Biology"]

---

Abstract: The human fossil record is one of the most complete for any mammal. A basal ancestral species, Australopithecus afarensis, exhibits a well-preserved postcranium that permits reconstruction of important events in the evolution of our locomotor skeleton. When compared to those of living apes and humans, it provides insights into the origin and design of the modern human frame. Evolutionary aspects of the human hip and thigh are reviewed, including the unusual corticotrabecular structure of the human proximal femur, and our markedly elongated lower limb. It is postulated that the latter may be more related to birthing capacity than to locomotion.
Categories: ["Natural History"]

---

Abstract: <user's input abstract goes here>
Reasoning: Let's think step by step in order to
\end{lstlisting}
\end{tcolorbox}
\caption{Few-shot examples for MeSH Biology classification (part 3)}
\label{fig:mesh-classification-prompt-part3}
\end{figure}

\clearpage

\begin{figure}[!htbp]
\centering
\begin{tcolorbox}[
  enhanced,
  colback=blue!10,
  colframe=blue!40,
  boxrule=0.6pt,
  arc=6pt,
  left=6pt,
  right=6pt,
  top=4pt,
  bottom=4pt,
  width=0.95\linewidth,
  title={\small\textbf{User}},fonttitle=\bfseries
]
\begin{lstlisting}[basicstyle=\small\rmfamily,breaklines=true,breakatwhitespace=true,columns=fullflexible]
You are a scientific classifier trained to match PubMed's MeSH indexing patterns. 
Given a question-answer pair from a biology research paper, classify it into one or more 
of the following 21 MeSH Biology subcategories.

IMPORTANT: Be CONSERVATIVE in your assignments. Most texts should have 1-2 
categories. Only assign a category if the text's MAIN FOCUS is clearly that category.

MeSH Biology Categories:
1. Botany - Study of plants (not just mentioning plants)
2. Cell Biology - Study of cell structure/function (not just using cells as models)
3. Computational Biology - Use of computational methods in biology
4. Cryobiology - Study of effects of low temperatures on living things
5. Cytology - Study of cells (microscopic anatomy)
6. Developmental Biology - Study of growth and development of organisms
7. Ecology - Study of organisms and their environment interactions
8. Exobiology - Study of life beyond Earth / astrobiology
9. Genetics - Study of genes, heredity, and genetic variation
10. Laboratory Animal Science - Study of lab animals for research
11. Marine Biology - Study of ocean organisms and ecosystems
12. Microbiology - Study of microorganisms (bacteria, viruses, fungi, etc.)
13. Molecular Biology - Molecular mechanisms (DNA/RNA/protein synthesis) - NOT general biochemistry
14. Natural History - Observational study of organisms in nature
15. Neurobiology - Study of nervous system biology
16. Parasitology - Study of parasites and parasitic diseases
17. Photobiology - Study of effects of light on living organisms
18. Radiobiology - Study of effects of radiation on living organisms
19. Sociobiology - Study of social behavior in biological terms
20. Synthetic Biology - Design and construction of new biological entities
21. Zoology - Study of animals

{few_shot_examples}

Question: {question}

Answer: {answer}

Respond with ONLY a JSON array of category names that apply to this QA pair.
Example response: ["Microbiology", "Genetics"]

If none of the categories apply well, respond with an empty array: []

Categories:
\end{lstlisting}
\end{tcolorbox}
\caption{Prompt template for QA MeSH classification}
\label{fig:mesh-qa-classification-prompt}
\end{figure}

\clearpage

\subsection{QA generation prompts}\label{sec:qa-generation-prompts}

Prompts used for generating, grading, and filtering reasoning problems. Prompts in Figures~\ref{fig:free-form-extractor-prompt-1},~\ref{fig:free-form-extractor-prompt-2}, and~\ref{fig:free-form-grader-prompt} are partially inspired by the prompts used in \citet{naturalreasoning} as well as in \citet{megascience}. \

\FloatBarrier

\begin{figure}[!ht]
\centering
\begin{tcolorbox}[enhanced,colback=gray!15,colframe=gray!50,boxrule=0.6pt,arc=6pt,
  left=6pt,right=6pt,top=4pt,bottom=4pt,width=0.95\linewidth,
  title={\small\textbf{System}},fonttitle=\bfseries]
\small
You are an expert content evaluator who determines if text content is relevant to the core scientific/technical content of a paper versus non-relevant material like copyright notices, licensing information, references, acknowledgments, or metadata.
\end{tcolorbox}
\vspace{2pt}
\begin{tcolorbox}[enhanced,colback=blue!10,colframe=blue!40,boxrule=0.6pt,arc=6pt,
  left=6pt,right=6pt,top=4pt,bottom=4pt,width=0.95\linewidth,
  title={\small\textbf{User}},fonttitle=\bfseries]
\small
Evaluate the following text chunk and determine if it contains core scientific/technical content that would be appropriate for generating educational questions. \\

TEXT CHUNK:

\{chunk\_text\}\\

EVALUATION CRITERIA:\\

- CORE CONTENT (High relevance): Scientific concepts, research findings, technical explanations, methodology, data analysis, theories, experimental results, clinical information, etc.\\
- NON-CORE CONTENT (Low relevance): Copyright notices, licensing text, reference lists, acknowledgments, author information, publication metadata, figure/table captions only, page headers/footers, disclaimers, etc. \\

SCORING:

- Score 8-10: Rich core content ideal for question generation \\
- Score 5-7: Some core content but mixed with non-relevant material \\
- Score 1-4: Primarily non-relevant content (references, metadata, etc.) \\

Provide your response in this format: \\

RELEVANCE\_SCORE: \textless numeric score between 1-10\textgreater \\

REASONING: \textless brief explanation of why this content is or isn't relevant for question generation\textgreater \\

CONTENT\_TYPE: \textless primary type of content: `core\_scientific', `mixed', `references', `metadata', `copyright', etc.\textgreater \\
\end{tcolorbox}
\caption{Prompt template for content relevance evaluation}
\label{fig:content-relevance-prompt}
\end{figure}

\clearpage

\FloatBarrier

\begin{figure}[!htp]
\centering
\begin{tcolorbox}[enhanced,colback=gray!15,colframe=gray!50,boxrule=0.6pt,arc=6pt,
  left=3pt,right=3pt,top=4pt,bottom=4pt,width=\textwidth,
  title={\small\textbf{System}},fonttitle=\bfseries]
\small
You are a helpful assistant that generates high-quality multiple-choice questions based on text provided by the user. Each question should be challenging but fair, with one clearly correct answer and plausible but incorrect distractors.
\end{tcolorbox}
\vspace{2pt}
\begin{tcolorbox}[enhanced,colback=blue!10,colframe=blue!40,boxrule=0.6pt,arc=6pt,
  left=3pt,right=3pt,top=4pt,bottom=4pt,width=\textwidth,
  title={\small\textbf{User}},fonttitle=\bfseries]
\small
Generate ONE well-formed multiple-choice question with exactly \{num\_answers\} answer choices labeled 1 through \{num\_answers\}. \\

Text:

\{text\_chunk\} \\

Requirements:

1. Begin with 1-2 sentences of contextual information that establishes the domain/topic without referencing source materials. \\
2. Create a challenging question that tests deep understanding. \\
3. Ensure there is EXACTLY ONE clearly correct answer. \\
4. Make the other choices plausible but clearly incorrect. \\
5. The question should focus on a concept or fact that is clearly stated or strongly implied in the text. \\
6. Number your answer choices from 1 to \{num\_answers\}. \\
7. IMPORTANT: Place the correct answer in position \{target\_correct\_position\}. The correct answer must be choice number \{target\_correct\_position\}. \\
8. DO NOT provide explanations for why each answer is correct or incorrect. \\
9. CRITICAL: Both context and question must be completely self-contained. DO NOT reference any external materials including: `the text', `the passage', `the document', `the paper', `the study', `the author states', `according to the text', `as mentioned', `as described', `Appendix', `Figure', `Table', `Section', `Chapter', `above', `below', or any other references to source materials or external content. \\
10. The context and question should read as if testing general knowledge on the topic, not comprehension of a specific text. \\
11. Answer choices should contain only direct technical information without meta-references to content or sources. \\

Your response must follow this format precisely: \\

CONTEXT: \textless 1-2 sentences establishing domain/topic context\textgreater \\

QUESTION: \textless the question\textgreater \\

1: \textless first answer choice\textgreater

2: \textless second answer choice\textgreater

... \\

CORRECT ANSWER: \{target\_correct\_position\}
\end{tcolorbox}
\caption{Prompt template for generating multiple-choice questions}
\label{fig:mcq-generation-prompt}
\end{figure}

\clearpage

\FloatBarrier

\begin{figure}[!htp]
\centering
\begin{tcolorbox}[enhanced,colback=gray!15,colframe=gray!50,boxrule=0.6pt,arc=6pt,
  left=3pt,right=3pt,top=4pt,bottom=4pt,width=\textwidth,
  title={\small\textbf{System}},fonttitle=\bfseries]
\small
You are an expert teacher evaluating the quality of a multiple choice question. Your role is to ensure questions are clear, fair, and educationally valuable.
\end{tcolorbox}
\vspace{2pt}
\begin{tcolorbox}[enhanced,colback=blue!10,colframe=blue!40,boxrule=0.6pt,arc=6pt,
  left=3pt,right=3pt,top=4pt,bottom=4pt,width=\textwidth,
  title={\small\textbf{User}},fonttitle=\bfseries]
\small
Evaluate the following multiple-choice question on a scale from 1-10, where 10 is a perfect question.\\

CONTENT:

\{chunk\_text\} \\

QUESTION:

\{question\} \\

CONTENT RELEVANCE INFO:

- Relevance Score: \{relevance\_check['relevance\_score']\}/10  \\
- Content Type: \{relevance\_check['content\_type']\}  \\
- Relevance Reasoning: \{relevance\_check['reasoning']\}  \\

Rate the question based on these criteria: \\

- Clarity: Is the question clear and unambiguous? \\
- Accuracy: Is the content factually correct and aligned with the source material? \\
- Difficulty: Is the difficulty appropriate (challenging but fair)? \\
- Distractors: Are the incorrect options plausible but clearly wrong? \\
- Educational value: Does answering this question demonstrate understanding? \\
- Self-contained: CRITICAL - Does the question stand alone without ANY references to external materials? \\
- Content relevance: IMPORTANT - Questions based on low-relevance content (references, metadata, etc.) should receive lower scores. \\

AUTOMATIC DISQUALIFIERS (score must be 1-3 if ANY are present):

- References to `the text', `the passage', `the document', `the paper', `the study' \\
- References to `the author', `according to', `as mentioned', `as described' \\
- References to `Appendix', `Figure', `Table', `Section', `Chapter' \\
- References to `above', `below', `previously mentioned', `following' \\
- Any other references that assume the reader has access to external materials \\
- Content based primarily on references, copyright notices, or metadata (should score 1-4)

SCORING ADJUSTMENT FOR CONTENT RELEVANCE:

- If content relevance score is 1-4: Maximum question score should be 4 \\
- If content relevance score is 5-7: Maximum question score should be 7 \\
- If content relevance score is 8-10: Normal scoring applies \\

A truly self-contained question should read like a general knowledge question on the topic. \\

Provide your response in this format: \\

SCORE: \textless numeric score between 1-10\textgreater \\

CRITIQUE: \textless brief explanation of score\textgreater \\
\end{tcolorbox}
\caption{Prompt template for evaluating multiple-choice question quality}
\label{fig:mcq-quality-evaluation-prompt}
\end{figure}

\clearpage

\begin{figure}[!ht]
\centering
\begin{tcolorbox}[enhanced,colback=gray!10,colframe=gray!40,boxrule=0.6pt,arc=6pt,
  left=6pt,right=6pt,top=4pt,bottom=4pt,width=0.95\linewidth,
  title={\small\textbf{System}},fonttitle=\bfseries]
\small
You are a factual and intelligent scientific assistant that identifies and extracts challenging, high-quality, exam-level question-and-answer pairs from text passages extracted from biology research documents.
\end{tcolorbox}

\vspace{4pt}

\begin{tcolorbox}[enhanced,colback=blue!10,colframe=blue!40,boxrule=0.6pt,arc=6pt,
  left=6pt,right=6pt,top=4pt,bottom=4pt,width=0.95\linewidth,
  title={\small\textbf{User}},fonttitle=\bfseries]
\small
Given a text passage from a scientific document in biology, determine if a challenging, high-quality, exam-level problem can be derived directly from the text. To do this, follow each step under the \#\#\# Instructions section below: \\

\#\#\# Instructions: \\

Step 1.) Evaluate the text passage on all of the bullet points for each criteria in the \#\#\# Text Evaluation Criteria section, keeping track of the total score.

Step 2.) Generate a question-and-answer pair by following ALL of the instruction steps in the \#\#\# Exam Question Instructions section.

Step 3.) Return your final response object according to the \#\#\# Response Instructions section. \\

\#\#\# Text Evaluation Criteria: \\

** Criteria 1 - Problem Completeness: **

- The text contains enough scientific content such that a question and corresponding answer can be derived directly from the text (+1 points) \\
- The text contains content that is appropriate for deriving an exam-level fill-in-the-blank problem (+1 points) \\
- The text minimizes non-scientific content, and does not rely on excessive references to external materials such as authors, tables, figures, or additional text that are not self-contained within the text passage (+1 points) \\

** Criteria 2 - Text Difficulty and Reasoning **

- The scientific content of the text passage is at least graduate-level or above (+1 points) \\
- The text passage contains sufficiently advanced scientific reasoning, technical explanations and/or discussions, computation, proofs, principles, or pathways (+1 points) \\
- The text contains content that is not just simple fact recall or concept recall (+1 points) \\

** Criteria 3 - Accuracy and Technical Correctness **

- The scientific content of the text does not contain any technical errors and is factually correct (+1 points) \\
- The text contains clear, concise, and accurate explanations that have sufficient technical and logical rigor (+1 points) \\
- Principles, theorems, concepts, facts, computations, or proofs critical for understanding the passage are included (+1 points) \\
\end{tcolorbox}
\caption{Prompt template for extracting free-form QA (part 1)}
\label{fig:free-form-extractor-prompt-1}
\end{figure}

\clearpage

\begin{figure}[!ht]
\centering
\begin{tcolorbox}[enhanced,colback=blue!10,colframe=blue!40,boxrule=0.6pt,arc=6pt,
  left=6pt,right=6pt,top=4pt,bottom=4pt,width=0.95\linewidth,
  title={\small\textbf{User (cont.)}},fonttitle=\bfseries]
\small
Follow ALL of the bullet points below when constructing your question-and-answer pair: \\

\#\#\# Exam Question Instructions:

1. If the text passage DOES NOT have a score of 9, it is NOT appropriate for deriving an exam-level problem. Skip to \#\#\# Response Instructions and return NONE for all fields. \\
2. CRITICAL: If you are unable to derive a question-and-answer pair that satisfies all of the steps below for ANY reason, skip to \#\#\# Response Instructions and return NONE for all fields. \\
3. Begin with 1-2 sentences of concise information that establishes the domain/topic WITHOUT referencing any source materials in the text passage. \\
4. Create a challenging, concise, exam-level question that requires extensive reasoning, sequential computation, sequential logic, or pathway logic rather than simple recall. It must be graduate-level or above in difficulty. \\
5. CRITICAL: Derive a question such that the corresponding answer is a single name, entity, or value that is easily verifiable. Questions with corresponding answers that are chemical compounds, protein names, quantitative values, pathways, or specific genes are appropriate. \\
6. CRITICAL: Corresponding answers must be a numerical value, name, or short entity STRICTLY. Answers that are calculations, statements, long entities, or explanations are NOT appropriate. \\
7. Ensure the question has a single clear, precise, and unambiguous answer. \\
8. Questions and their corresponding answers should be inspired from the content in the text, but CANNOT reference the text passage itself. \\
9. Questions may involve mechanisms, pathways, evolutionary principles, genetic analysis, experimental design, computational biology, biological theorems, or systems-level understanding. \\
10. CRITICAL: The context, question and answer must be completely self-contained. DO NOT reference any external materials including: `the text', `the passage', `the document', `the paper', `the study', `the author states', `according to the text', `Appendix', `Figure', `Table', `Section', `Chapter', `above', `below', or any other references to source materials. \\
11. The context and question should read like an exam question for graduate-level students and above, testing general knowledge or reasoning on biology, not comprehension of a specific text. \\
12. The answer must contain only direct technical information without any meta-references to content, studies, or source materials. \\
13. If your question requires an answer in a specific format, specify that in the question itself at the end. \\

\#\#\# Response Instructions: \\

Your response must follow this format precisely:

CONTEXT: \textless 1-2 sentences establishing domain/topic context; return NONE if problem cannot be derived\textgreater

QUESTION: \textless your exam-level question; return NONE if problem cannot be derived\textgreater

ANSWER: \textless your final answer wrapped in LaTeX boxed format \$\$\textbackslash boxed\{\{your\_final\_answer\}\}\$\$; return NONE if problem cannot be derived\textgreater

REASONING: \textless your reasoning for solving the question; return NONE if problem cannot be derived\textgreater \\

\#\#\# Text Passage:

\{augmented\_chunk\}
\end{tcolorbox}
\caption{Prompt template for extracting free-form QA (part 2)}
\label{fig:free-form-extractor-prompt-2}
\end{figure}

\clearpage

\begin{figure}[!htp]
\centering
\begin{tcolorbox}[enhanced,colback=gray!10,colframe=gray!40,boxrule=0.6pt,arc=6pt,
  left=6pt,right=6pt,top=4pt,bottom=4pt,width=0.95\linewidth,
  title={\small\textbf{System}},fonttitle=\bfseries]
\small
You are a factual and intelligent biology exam grader that will assess the quality of a question-and-answer pair and determine if it is a challenging, high-quality, and exam-level problem appropriate for biology.
\end{tcolorbox}

\vspace{4pt}

\begin{tcolorbox}[enhanced,colback=blue!10,colframe=blue!40,boxrule=0.6pt,arc=6pt,
  left=6pt,right=6pt,top=4pt,bottom=4pt,width=0.95\linewidth,
  title={\small\textbf{User}},fonttitle=\bfseries]
\small
Your task is to verify the quality of a biology question-and-answer pair, given under the \#\#\# EXAM PROBLEM section. \\
To verify the exam problem, follow the instructions below. \\

\#\#\# Grading Instructions

Score the question-and-answer pair based on the \#\#\# Scoring Criteria section. If the pair is NONE, skip to the \#\#\# Response Instructions. \\

\#\#\# Scoring Criteria: \\

A valid question-and-answer pair must meet ALL of the following criteria: \\

* The question should contain a problem to be solved, instead of only presenting statements (+1 points) \\
* The question should be well-defined and self-contained, ie have all necessary information to derive an answer. It does not reference external materials without first providing a self-contained summary (+1 points) \\
* The question should be specific and clear. There should be one correct answer to the question (+1 points) \\
* The question should not refer to external resources such as figures, videos, etc. (+1 points) \\
* To derive an answer, multi-step reasoning, computations, proofs, intense logic, or recalling complex relevant knowledge is required (+1 points) \\
* The difficulty should be graduate-level or above (+1 points) \\
* The question can contain LaTeX but it must be correct (+1 points) \\
* The question should be worded such that the corresponding answer is a single name, entity, or value that is easily verifiable. Chemical compounds, protein names, quantitative values, pathways, or specific genes are appropriate (+1 points) \\
* Corresponding answers must be a numerical value, name, or short entity STRICTLY. Calculations, statements, long entities, or explanations are NOT appropriate (+1 points) \\

\#\#\# Response Instructions: \\

SCORE: \textless how well the pair meets all criteria using the total score\textgreater

CRITIQUE: \textless your critique of the question-and-answer pair\textgreater \\

\#\#\# EXAM PROBLEM

\{exam\_problem\}
\end{tcolorbox}
\caption{Prompt template for grading free-form QA}
\label{fig:free-form-grader-prompt}
\end{figure}

\clearpage

\subsection{Evaluation prompts}

\begin{figure}[!htbp]
\centering
\begin{tcolorbox}[enhanced,colback=blue!10,colframe=blue!40,boxrule=0.6pt,arc=6pt,
  left=6pt,right=6pt,top=4pt,bottom=4pt,width=0.95\linewidth,
  title={\small\textbf{User}},fonttitle=\bfseries]
\small
The following is a multiple choice question about biology.\\
Please answer by responding with the letter of the correct answer.\\[4pt]
Question: \{question\}\\[4pt]
Options:\\
\{answers\}\\[4pt]
You MUST include the letter of the correct answer within the following tags:\\
{[}ANSWER{]} and {[}/ANSWER{]}. For example, `{[}ANSWER{]}\textless answer\textgreater{[}/ANSWER{]}',\\
where \textless answer\textgreater{} is the correct letter. Always answer in exactly this format\\
of a single letter between the two tags, even if you are unsure.\\
We require this because we use automatic parsing.
\end{tcolorbox}
\caption{Prompt for LAB-Bench evaluation}
\label{fig:labbench-eval-prompt}
\end{figure}

\begin{figure}[!htp]
\centering
\begin{tcolorbox}[enhanced,colback=gray!15,colframe=gray!50,boxrule=0.6pt,arc=6pt,
  left=6pt,right=6pt,top=4pt,bottom=4pt,width=0.95\linewidth,
  title={\small\textbf{System}},fonttitle=\bfseries]
\small
You are an expert scientist with graduate-level knowledge in biology,
physics, and chemistry. Answer the following multiple-choice question.
First state your chosen answer (the letter and the corresponding option
text), then briefly explain your reasoning.
\end{tcolorbox}
\vspace{2pt}
\begin{tcolorbox}[enhanced,colback=blue!10,colframe=blue!40,boxrule=0.6pt,arc=6pt,
  left=6pt,right=6pt,top=4pt,bottom=4pt,width=0.95\linewidth,
  title={\small\textbf{User}},fonttitle=\bfseries]
\small
\{question\}\\[4pt]
A. \{option\_a\}\\
B. \{option\_b\}\\
C. \{option\_c\}\\
D. \{option\_d\}\\[4pt]
Think step by step, and return your final answer in \textbackslash boxed\{\}
\end{tcolorbox}
\caption{Prompt for evaluating GPQA-Diamond}
\label{fig:gpqa-eval-prompt}
\end{figure}

\begin{figure}[!ht]
\centering
\begin{tcolorbox}[enhanced,colback=gray!15,colframe=gray!50,boxrule=0.6pt,arc=6pt,
  left=6pt,right=6pt,top=4pt,bottom=4pt,width=0.95\linewidth,
  title={\small\textbf{System}},fonttitle=\bfseries]
\small
You are an expert biomedical researcher. You will be given context from a
research paper abstract and a yes/no/maybe question about the findings.
First state your chosen answer (the letter and the corresponding option
text), then briefly explain your reasoning based on the provided context.
\end{tcolorbox}
\vspace{2pt}
\begin{tcolorbox}[enhanced,colback=blue!10,colframe=blue!40,boxrule=0.6pt,arc=6pt,
  left=6pt,right=6pt,top=4pt,bottom=4pt,width=0.95\linewidth,
  title={\small\textbf{User}},fonttitle=\bfseries]
\small
Context:\\
\{abstract\_text\}\\[4pt]
Question: \{question\}\\[4pt]
A. yes\\
B. no\\
C. maybe\\[4pt]
Think step by step, and return your final answer in \textbackslash boxed\{\}
\end{tcolorbox}
\caption{Prompt for evaluating PubMedQA}
\label{fig:pubmedqa-eval-prompt}
\end{figure}

\clearpage

\subsection{Example reasoning question}
\FloatBarrier

\begin{figure}[!ht]
\centering
\begin{tcolorbox}[enhanced,colback=blue!10,colframe=blue!40,boxrule=0.6pt,arc=6pt,
  left=6pt,right=6pt,top=4pt,bottom=4pt,width=0.95\linewidth,
  title={\small\textbf{User}},fonttitle=\bfseries]
\small
Question:\\
Mathematical models are vital for understanding how\\
viruses compete within a host, especially when multiple\\
strains are present and subject to antiviral therapy.\\
The interplay between viral life cycle parameters and drug\\
regimen adherence can strongly influence infection outcomes.\\[4pt]
When simulating multi-strain viral competition under\\
imperfect therapy adherence, which factor directly increases\\
both the burst size and infectivity of a viral strain\\
but also heightens its risk of cell death before maturation?\\[4pt]
A.) Increasing the rate of precursor decay\\
B.) Reducing the dosing period for antiviral drugs\\
C.) Shortening the immature phase duration\\
D.) Lengthening the immature phase duration\\
E.) Decreasing the number of discrete maturation steps\\
F.) Lowering the probability of taking each scheduled dose\\
G.) Raising the antiviral drug efficacy
\end{tcolorbox}
\caption{Example reasoning question}
\label{fig:argonium-question}
\end{figure}

\end{document}